\documentclass[unnumsec,webpdf,contemporary,large]{oup-authoring-template}%

\graphicspath{{figures/}}

\usepackage{hyperref}
\usepackage{url}
\usepackage{algorithm}
\usepackage[disable]{todonotes}
\usepackage{algpseudocode}
\algnewcommand\Input{\item[\textbf{Input:}]}%
\algnewcommand\Output{\item[\textbf{Output:}]}%
\usepackage{graphicx}
\usepackage[capitalise,noabbrev,nameinlink]{cleveref}

\usepackage{booktabs}
\usepackage{multirow}
\usepackage{todonotes}
\usepackage{wrapfig}
\usepackage{lipsum}
\usepackage[dvipsnames]{xcolor}
\usepackage{framed}
\usepackage{array}
\usepackage{comment}
\usepackage{tabularx}
\usepackage{svg}
\svgsetup{inkscapepath=svgsubdir}
\usepackage{float}
\usepackage{amssymb}
\usepackage{bibunits}
\defaultbibliography{reference}
\defaultbibliographystyle{abbrvnat}

\usepackage{amsmath,amsfonts,bm}

\def\eqref#1{equation~\ref{#1}}

\def\1{\bm{1}}

\def\mA{{\bm{A}}}

\def\mK{{\bm{K}}}

\def\mQ{{\bm{Q}}}

\def\mV{{\bm{V}}}

\DeclareMathAlphabet{\mathsfit}{\encodingdefault}{\sfdefault}{m}{sl}
\SetMathAlphabet{\mathsfit}{bold}{\encodingdefault}{\sfdefault}{bx}{n}

\newcommand{\dataset}[0]{\mathcal{D}}

\theoremstyle{thmstyleone}%
\theoremstyle{thmstyletwo}%
\theoremstyle{thmstylethree}%

\begin{document}

\journaltitle{Journal Title Here}
\DOI{DOI added during production}
\copyrightyear{YEAR}
\pubyear{YEAR}
\vol{XX}
\issue{x}
\access{Published: Date added during production}
\appnotes{Paper}

\firstpage{1}


\title[TabPFN-Wide]{TabPFN-Wide: Continued Pre-Training \\ for Extreme Feature Counts}

\author[1, $\ast$]{Christopher Kolberg}
\author[1, $\ast$]{Jules Kreuer\ORCID{0000-0002-7305-833X}}
\author[1, $\ast$]{Jonas Huurdeman}
\author[1]{Sofiane Ouaari}
\author[2,3]{Katharina Eggensperger}
\author[1,$+$]{Nico Pfeifer\ORCID{0000-0002-4647-8566}}

\address[1]{\orgdiv{Institute for Bioinformatics and Medical Informatics}, \orgname{University of T\"{u}bingen}, \orgaddress{\street{Maria-von-Linden-Str. 6}, \postcode{72076}, \state{Baden-W\"{u}rttemberg}, \country{Germany}}}
\address[2]{\orgdiv{AutoML for Science}, \orgname{University of T\"{u}bingen}, \orgaddress{\street{Maria-von-Linden-Str. 6}, \postcode{72076}, \state{Baden-W\"{u}rttemberg}, \country{Germany}}}
\address[3]{\orgdiv{Lamarr Institute for Machine Learning and Artificial Intelligence}, \orgname{TU Dortmund}, \street{Joseph-von-Fraunhofer-Straße 25} \postcode{44227}, \country{Germany}}

\corresp[$\ast$]{These authors contributed equally\,;\, }
\corresp[$+$]{corresponding author: \href{email:nico.pfeifer@uni-tuebingen.de}{nico.pfeifer@uni-tuebingen.de}}

\received{Date}{0}{Year}
\revised{Date}{0}{Year}
\accepted{Date}{0}{Year}



\abstract{Revealing novel insights from the relationship between molecular measurements and pathology remains a very impactful application of machine learning in biomedicine. Data in this domain typically contain only a few observations but thousands of potentially noisy features, posing challenges for conventional tabular machine learning approaches. While prior-data fitted networks emerge as foundation models for predictive tabular data tasks, they are currently not suited to handle large feature counts ($>500$). Although feature reduction enables their application, it hinders feature importance analysis. We propose a strategy that extends existing models through continued pre-training on synthetic data sampled from a customized prior. The resulting model, TabPFN-Wide\footnote{Training code, package and model weights are released on \url{https://github.com/not-a-feature/TabPFN-Wide}.}, matches or exceeds its base model's performance, while exhibiting improved robustness to noise. It seamlessly scales beyond $30{,}000$ categorical and continuous features, regardless of noise levels, while maintaining inherent interpretability, which is critical for biomedical applications. Our results demonstrate that prior-informed adaptation is suitable to enhance the capability of foundation models for high-dimensional data. On real-world omics datasets, we show that many of the most relevant features identified by the model overlap with previous biological findings, while others propose potential starting points for future studies.
}


\keywords[Abbreviations]{HDLSS: High-Dimensional Low-Sample Size}

\maketitle

\section{Introduction}
\label{introduction}



Tabular data are an important data modality used for quantitative research in healthcare, finance, natural sciences, and many more. Tabular data are relevant for various real-world applications and offer ``uniquely exciting, large, unsolved challenges for researchers"~\citep{van2024tabular}. One main challenge is high-dimensional, low-sample-size (HDLSS) data, common in biomedical applications. Cohort sizes of studies are small due to cost, time, or disease rarity, while modern biomedical technologies, on the other hand, enable the measurement of thousands of features per patient. Collected data can then be examined for predictive tasks, for instance, to study interactions between thousands of biomarkers and cancer types~\citep{tcga-gbm,tcga-ov}. Even more importantly, to guide scientific discovery~\citep{CMKN,COmic}, interpretability is as important as accuracy. Such extreme feature counts in combination with a low sample size raise a difficulty for real-world machine learning applications. 

 
Foundation models for structured data have emerged, and models like TabPFN and TabICL~\citep{hollmann-arxiv23a,hollmann-nature25a-fix,qu2025tabicl} are currently at the forefront of predictive tabular ML benchmark tasks~\citep{erickson2025tabarena}. These models use in-context learning (ICL)~\citep{brown2020language} and are based on transformers ~\citep{vaswani2017attention}, pre-trained on synthetic or real-world data to solve tabular regression and classification tasks. As a result, they are highly effective on unseen tasks with characteristics similar to those seen during pre-training. 
While the exact training data are often unknown, empirical performance on HDLSS data suggests that current models have not learned to handle extreme feature counts.

Such limits stem from insufficient exposure during pre-training rather than a lack of model capacity, data, or resources; thus, retraining from scratch using a broader prior could be a solution. However, re-training from scratch whenever we encounter a new task or data characteristic to improve a model's performance would be extremely resource-intensive and, therefore, often infeasible. It would also contradict the concept of a foundation model, which is pre-trained to serve as the basis for downstream tasks. Naive solutions, such as subsampling or compressing features to match the dimensionality of the pre-training data, render methods for quantifying feature importance ineffective. Instead, we aim to enhance the capability of already existing pre-trained models in a resource-efficient way, while keeping the interpretability workflow intact. Concretely, we study the more general question: ``Can continued pre-training extend tabular foundation models to generalize across diverse task types in high-dimensional, small-sample data?''

To address these constraints, we propose TabPFN-Wide, a model built upon TabPFNv2 that seamlessly scales to large feature counts, thereby handling HDLSS data in biomedicine. 

Specifically, our contributions are:
\begin{enumerate}
    \item We develop a novel prior to efficiently generate synthetic HDLSS data.
    \item We propose continued pre-training to extend TabPFNv2, resulting in TabPFN-Wide, to handle extreme feature counts beyond $30{,}000$ features.
    \item In empirical evaluations on biomedical data and standard tabular benchmark tasks, we show that TabPFN-Wide maintains performance within its original range, while being significantly more robust on wide data.
    \item Finally, we study the inherent interpretability of attention maps of TabPFN-Wide and show that attention maps allow us to identify relevant features.
\end{enumerate}

\section{Materials and Methods}

\subsection{Problem Description}\label{ProblemDescription}
We start by briefly describing our problem setup and the challenges for robustly scaling tabular foundation models, specifically TabPFNv2~\citep{hollmann-nature25a-fix}, to thousands of features. 

\textbf{Tabular data} can be described as a dataset $\dataset = \{ ( x_i, y_i) \}_{i=1}^{n}$ containing $n$ samples (rows). Each sample consists of a feature vector $x_i \in \mathbb{R}^m$ with $m$ features (columns) and, for classification tasks, a corresponding label $y_i \in \{1, 2, ..., C\}$.
To measure performance of a model $f$, we split available data into a train dataset $\dataset_{train} = \{ (x_i^{(train)}, y_i^{(train)}) \}_{i=1}^{n_{train}}$ and a validation dataset $\dataset_{val} = \dataset \smallsetminus \dataset_{train}$ and compute a loss, e.g., log loss, $\mathcal{L} = \sum_{\left(x_i, y_i\right) \in \dataset_{val}}l\left(f(x_i, \dataset_{train}), y_i\right)$ to approximate how well $f$ would generalize to unseen (test) samples.
What distinguishes tabular data from other modalities are their heterogeneous feature types (categorical, numerical, missing values), and potentially diverse structures with the number of samples and features ranging from a few to millions~\citep{van2024tabular}.

\textbf{HDLSS data} are a specific type of tabular data where the number of features is much larger than the number of samples, i.e., $m \gg n$. Such data typically occur in the biomedical domain. For example, cancer data from The Cancer Genome Atlas (TCGA) provide high-dimensional multi-omics measurements from cancer patients, such as those with ovarian cancer~\citep{tcga-ov}. In this setting, a typical classification problem is the identification of cancer subtypes. Improving the accuracy and robustness of predictive machine learning models supports precise diagnoses and personalized treatments, ultimately improving patient outcomes. A key difficulty arises from the high-dimensional feature space of molecular data, where noisy or irrelevant measurements often obscure subtype-specific signals. This complexity inhibits the detection of biologically meaningful patterns and hinders the ability to distinguish molecular differences between tumor subtypes.

\textbf{Biomedical downstream tasks demand interpretability} due to their sensitive nature. However, for HDLSS data, common post-hoc interpretability methods are unreliable~\citep{posthoc-fail}. For example, traditional permutation-based testing approaches like SHAP~\citep{lundberg2017unified} require computing scores for each variable multiple times across multiple permutations, making it computationally demanding for high-dimensional datasets. Additionally, the low sample size reduces the stability of the results.

Consequently, feature reduction or selection techniques are applied beforehand to reduce the number of features to a computable range. Yet, this inherently poses the risk of losing information or dropping potentially relevant features, which would be highly undesirable for applications in the real world. Thus, we avoid feature reduction and instead make our model work on all available features. This allows the model to identify the most predictive features directly. To gain insights into this internal selection process, we sought inherent interpretability methods and chose to use attention maps computed within the transformer architecture.
However, the role and interpretability of attention maps are controversial in the literature, with nearly no previous work on attention analysis of TabPFN (or related models). In the context of large language models (LLMs), studies have shown that while attention maps may provide a coarse indication of a model's reasoning process, they are often noisy and can erroneously emphasize irrelevant tokens~\citep{serrano-smith-2019-attention, Jain2019AttentionIN}. Nevertheless, there have been successful approaches in biomedicine, where features identified by studying attention maps overlap with biological knowledge~\citep{CMKN,COmic}.

For TabPFNv2's attention specifically, earlier research shows that it evolves across layers, shifting from label-focused attention in the first layers to semantically relevant attribute attention in deeper layers~\citep{ye2025closerlooktabpfnv2}. Additionally, \citet{rubachev2025finetuningtabularfoundationmodels} links a reduced entropy of the attention score distribution to a more focused classification model. Building on these observations, we examine the attention maps as described, with careful consideration of their potential shortcomings. 

\begin{figure*}[htbp]
    \begin{minipage}{0.5\textwidth}
        \centering
        \begin{algorithm}[H]
        \caption{Continuous Feature Widening}

        \begin{algorithmic}[1]
        \small
        \Input \colorbox{blue!30}{Input features ${X_{\text{cont}}} \in \mathbb{R}^{n \times m_{\text{cont}}}$},
            \colorbox{orange!30}{target dimension $d_{\text{cont}}$}, \newline
            \colorbox{green!30}{sparsity~$p~\in~[0,1]$},
            \colorbox{magenta!30}{noise std. $\sigma$} 
        \Output \colorbox{cyan!30}{Wide continuous features $X_{\text{wide\_cont}} \in \mathbb{R}^{n \times d}$}
    
        \State Sample weights $W \in \mathbb{R}^{m \times \colorbox{orange!30}{$d_{\text{cont}}$}}$ with $W_{ij} \sim \mathcal{N}(0,1)$
        \State Sample mask $M \in \{0,1\}^{m \times \colorbox{orange!30}{$d_{\text{cont}}$}}$ with $M_{ij} \sim \mathrm{Bernoulli}(\colorbox{green!30}{p})$
        \State Compute wide features \colorbox{cyan!30}{$X_{\text{wide\_cont}}$}$\gets$ \colorbox{blue!30}{$X_{\text{cont}}$} $(M \odot W)$
        \State Sample noise $N \in \mathbb{R}^{m \times \colorbox{orange!30}{$d_{\text{cont}}$}}$ \newline
        \hspace*{1.5em} with $N_{ij} \sim \mathcal{N}(0,$ \colorbox{magenta!30}{$\sigma$}$\sigma_j)$ and 
        $\sigma_j=\mathrm{std}($\colorbox{cyan!30}{$X_{\text{wide\_cont}_{:,j}}$}$)$
        \State Add noise \colorbox{cyan!30}{$X_{\text{wide\_cont}}$}$ \gets $\colorbox{cyan!30}{$X_{\text{wide\_cont}}$}$ + N$
        \State \Return \colorbox{cyan!30}{$X_{\text{wide\_cont}}$}
        \end{algorithmic}
        \label{alg:continuous-widening}
        \end{algorithm}
    \end{minipage}
    \hfill
    \begin{minipage}{0.46\textwidth}
    \centering
        \begin{algorithm}[H]
\caption{Categorical Feature Widening}
\label{alg:categorical-widening}
\begin{algorithmic}[1]
\small
\Input \colorbox{blue!30}{Input features $X_{\text{cat}} \in \mathbb{R}^{n \times m_{\text{cat}}}$},
       \colorbox{orange!30}{target dimension $d_{\text{cat}}$}, \newline
       \colorbox{green!30}{sparsity $p$},
       \colorbox{magenta!30}{max categories $K_{\max}$}
\Output \colorbox{cyan!30}{Wide categorical features $X_{\text{wide\_cat}} \in \mathbb{R}^{n \times d_{\text{cat}}}$}

\For{$j = 1$ \textbf{to} \colorbox{orange!30}{$d_{\text{cat}}$}}
    \State $k \gets \max(1, \lfloor \colorbox{green!30}{p} \cdot m_{\text{cat}} \rfloor)$
    \State Sample donor column indices $\mathcal{D} \subseteq \{1,\dots,m_{\text{cat}}\}$ with $|\mathcal{D}| = k$
    \For{$i = 1$ \textbf{to} $n$}
        \State Sample donor column index $d_s \in \mathcal{D}$ using the uniform distribution
        \State \colorbox{cyan!30}{$X_{\text{wide\_cat}}[i,j]$} $\gets \colorbox{blue!30}{$X_{\text{cat}}$}[i,d_s]$
    \EndFor
    \State Sample $K_j \in \{3,\dots,$\colorbox{magenta!30}{$K_{\max}$}$\}$ using the uniform distribution
    \While{$\mathrm{unique}$(\colorbox{cyan!30}{$X_{\text{wide\_cat}}$}$[:,j]) > K_j$}
        \State Merge the rarest category into one of the $K_j$ most frequent 
        \State categories randomly using the uniform distribution

    \EndWhile
\EndFor

\State \Return \colorbox{cyan!30}{$X_{\text{wide\_cat}}$}
\end{algorithmic}
\end{algorithm}        
    \end{minipage}
    
    \caption{Pseudocode of continuous widening (left) and categorical widening (right).}
\end{figure*}

\subsection{Tabular Foundation Models for Predictive ML Tasks}
\label{sec:tab_background}

\textbf{Prevailing models changed from traditional to pre-trained models.} 
Traditional ML models, like random forests or multi-layer perceptrons, must be trained from scratch for each task, with their predictive quality depending on hyperparameters and encoded inductive biases. With the rise of transformer models, amortized inference as a new learning paradigm for tabular data has emerged. Such foundation models are trained across many (synthetic) tasks to \textit{learn how to do statistical inference} via ICL. 
At inference time, training samples and query points are fed to the model, which then approximates Bayesian inference to predict labels~\citep{muller-iclr22a, muller-position-2025}.

The use of ICL for predictive tabular tasks was originally based on LLMs. Further building on the successes of LLMs, numerous studies have investigated their application to tabular data \citep{hegselmann2023tabllm,zhang2024tablellamaopenlargegeneralist,Herzig_2020}. For these approaches, natural language representations of the tables are used for few- and zero-shot tabular classification. 
However, table-to-text-based models are limited by the context window of the underlying LLM; their predictions could be based on learned world knowledge rather than the table data, and, importantly, they cannot inherently leverage the structure (columns and rows) of tabular data. While yielding impressive results for zero- and few-shot tasks, they perform worse when more data are available~\citep{hegselmann2023tabllm}. 
To address these weaknesses, while simultaneously keeping the ICL approach, tabular foundation models emerged, with TabPFN \citep{hollmann2022tabpfn} being one of the earliest representatives.
It is entirely trained on synthetic data generated from a prior based on structural causal models, yielding competitive performance on unseen tabular classification tasks. TabPFNv2~\citep{hollmann-nature25a-fix}, a follow-up, introduced a modified prior and architecture, achieving state-of-the-art performance on datasets with up to $10{,}000$ samples and $500$ features.

\textbf{Current research focuses on extending the applicability regarding the number of samples and computational cost.} One prominent example is TabICL \citep{qu2025tabicl}, which uses only a fixed number of embedded [CLS] tokens per sample for ICL rather than all the features. Furthermore, TuneTables~\citep{feuer2024tunetables} optimizes the context of TabPFN using a learned compact dataset representation instead of the whole training data. Additionally, TabFlex~\citep{tabflex} uses linear attention instead of standard (quadratic) attention to reduce complexity. Other research directions focus on localization approaches to select relevant context samples~\citep{tabdpt,mixturepfn,koshil-2024}. While all these approaches aim to extend the application range, they propose new architectures and inference mechanisms, often applying feature reduction and compression. In contrast, we aim to expand an \textit{existing} model's capability without impairing interpretability on a per-feature level.
For these reasons, we focus on TabPFNv2~\citep{hollmann-nature25a-fix}, currently the only state-of-the-art approach that can simply be modified (see \cref{subsec:att_maps}) to satisfy our requirement of preserving a per-feature resolution throughout its architecture. 

\textbf{Fine-tuning and continued pre-training improve performance on downstream tasks.}
Fine-tuning, i.e., performing gradient updates using data from the target downstream tasks, is commonly used to adapt LLMs to application domains~\citep{christophe2024med,weyssow2024exploringpeft} and has been proposed as a best practice to compare models~\citep{zhang-trainbeforetest-2025}. Similarly, fine-tuning TabPFN in general~\citep{forestpfn,rubachev2025finetuningtabularfoundationmodels} or specifically performing parameter-efficient fine-tuning for context optimization~\citep{feuer2024tunetables} can improve performance on a single downstream task. However, this requires a sufficient number of samples for this task.
Continued pre-training, in contrast, does not use data from the target task but leverages tasks with properties similar to the target task. For example, Real-TabPFN~\citep{garg2025realtabpfnimprovingtabularfoundation}, further pre-trained on real-world datasets, shows significant improvements on real-world tabular benchmarks. We follow this direction, but instead of using real-world data, we study how to continue pre-training with synthetic data to scale TabPFN to extreme feature counts, far beyond what it has seen during pre-training.
Because this involves sequential training, it is crucial to prevent the model from experiencing catastrophic forgetting~\citep{catastrophic_forgetting,measure_catastrophic_forgetting}. This could cause the model to perform significantly worse on tabular data within the original ranges of TabPFNv2.

\subsection{Methodology}
\label{sec:methodology}

We propose a novel approach to extend the capabilities of tabular foundation models, specifically TabPFNv2, while preserving per-feature interpretability. We split our method into three components: First, we develop a prior to efficiently generate synthetic HDLSS data. Second, we use this data to continue pre-training, and third, we study attention maps for feature-wise interpretability. 

\subsubsection{A Prior for Synthetic HDLSS Data Generation}

To adapt our model, we need a mechanism to generate training data, which (1) works fast and cost-effectively, since we need multiple datasets per batch step, and (2) yields realistic data to provide a meaningful and reliable signal during adaptation. 

\noindent\textbf{HDLSS prior.} For the first desideratum, we follow prior work and rely on synthetic data obtained from a data-generating mechanism based on structural causal models~\citep{hollmann2022tabpfn,hollmann-arxiv23a}. Datasets are therefore drawn from randomly sampled directed acyclic graphs. Specifically, as the TabPFNv2 prior is not publicly available, we use the open-source prior used to train TabICL~\citep{qu2025tabicl}, considering TabICL's strong empirical performance as evidence of the prior's similar effectiveness.
To satisfy the second desideratum, we exploit the observation that features in HDLSS datasets typically exhibit substantial noise and strong inter-feature correlations~\citep{HDLSS_properties}. 
 

Based on this assumption, we construct a feature widening prior that can widen continuous features as formalized in \cref{alg:continuous-widening} as well as categorical features as shown in \cref{alg:categorical-widening}.
During training, we first sample a dataset with a moderate number of features $m$ from the TabICL prior and subsequently widen it to a target dimension $d \gg m$. Since datasets within a batch do not necessarily share the same feature semantics, feature widening is applied independently per dataset. 
The widening procedure distinguishes between continuous and categorical features and allocates the target dimensionality accordingly. To this end, we first identify the feature types present in the dataset. A feature is considered categorical if it has at most $20$ distinct values; all remaining features are treated as continuous. Let $r_{\mathrm{cat}}$ denote the resulting categorical ratio, defined as the number of categorical features divided by the total number of features.
Given a target number of features to be added, we allocate
$d_{\mathrm{cat}} = \left\lfloor r_{\mathrm{cat}} \cdot (d - m)\right\rfloor$
categorical features and
$d_{\mathrm{cont}} = (d - m) - d_{\mathrm{cat}}$
continuous features.
Continuous features are widened as described in \cref{alg:continuous-widening}. Specifically, we sample a sparse linear transformation with sparsity $p$ (lines~1-2) and apply it to the original features to obtain new features (line~3). Feature-dependent Gaussian noise is then added to the projected features (lines~4-5), ensuring realistic variability while preserving correlation structure.


Categorical features are widened using the complementary mechanism in \cref{alg:categorical-widening}.  
New categorical features are generated by sparsely sampling dependencies on existing categorical features using the same sparsity parameter $p$ as in the continuous widening procedure (line~3) and copying feature values on a per-sample basis (lines~5-6). To prevent degenerate high-cardinality variables, each generated feature is constrained to a bounded number of categories via a category reduction step (line~10). The target cardinality of each feature is sampled from a discrete exponential distribution, biasing the process towards low-cardinality features while still allowing higher-cardinality cases.  
With this procedure, we can generate thousands of new features highly correlated to the original feature set, mimicking HDLSS data. 

\todo{Move to appendix}
Importantly, the sparsity parameter $p$ allows us to control the induced correlation patterns, matching the dense and sparse correlation structures observed in real-world biomedical data (see Appendix L for a detailed visual comparison).
\subsubsection{Continued Pre-Training}
\label{subsec:continued_pretraining}
For our continued pre-training setup, we start from the original TabPFNv2 classifier checkpoint\footnote{See \href{https://huggingface.co/Prior-Labs/TabPFN-v2-clf/resolve/main/tabpfn-v2-classifier.ckpt}{Hugging Face model}; Runtime complexity remains unaffected, thus, to satisfy higher resource demands for continued pre-training we used 4 NVIDIA H100 GPUs with a combined memory of 320GB.} and updated all parameters during training. 
We used AdamW \citep{loshchilov2017decoupled}  (using a weight decay of $1\times10^{-4}$ and a learning rate of $1\times10^{-5}$) with linear warm-up, cosine decay, and gradient norms clipping to $1.0$. We used a batch size of $16$, reducing it to $8$ for training runs with over $5{,}000$ features due to memory constraints. Training and validation were performed using cross-entropy loss. The generated datasets of the TabICL prior had up to $10$ classes (to match TabPFNv2's limitations), $40$ to $400$ samples, and $50$ to $350$ features, which we then widened using \cref{alg:continuous-widening} and \ref{alg:categorical-widening}. 
The target number of features $d$ in was uniformly sampled between $200$ and a predefined maximum $d_{max}$, with $d_{max} \in \{1{,}500, 5{,}000, 8{,}000\}$. We trained separate models for each $d_{max}$. 

With probability $0.5$, the original features were appended to the final dataset and afterwards the feature order was randomly permuted. Sparsity and noise level were uniformly sampled with $p \in [0, 0.05]$ and $\sigma \in [0, 1]$, following the analysis visualized in the Appendix L. 
We denote the resulting models as TabPFN-Wide{-*}, where * indicates the maximum number of features used during training. 

We fixed the total training duration to 10,000 optimization steps for all models, as validation ROC-AUC on a set of omics and synthetic SNP datasets plateaued beyond this point. These datasets were used exclusively to observe convergence rather than for active checkpoint selection, with further details provided in Appendix J.

%

\subsubsection{Feature-wise Interpretability via Attention Maps}
\label{subsec:att_maps}
To gain insights into TabPFNv2's inference, we analyze attention maps, focusing on attention towards the label as a proxy for feature importance. This requires that each transformer (token) column corresponds to a dataset feature. By default, TabPFNv2 groups features, adds distribution-dependent features, or may remove features impairing a token-to-feature mapping. To address this, we disabled these modifications for training as well as our biomedical datasets and interpretability analyses.
Attention maps are an intermediate step of the original dot-product attention computation \citep{vaswani2017attention} and we refer to the matrix $\mA$ in \cref{eq:attention} as ``attention map", with query matrix $\mQ$, key matrix $\mK$, value matrix $\mV$, and key vector dimensionality $d_{key}$:
\begin{equation}
    \text{Attention}(\mQ, \mK, \mV) = \text{softmax}\left(\frac{\mQ\mK^T}{\sqrt{d_{key}}}\right)\mV = \mA \mV.
    \label{eq:attention}
\end{equation}
To interpret attention maps as an indicator of feature importance, we consider only TabPFNv2's feature-wise attention, disregarding the sample-wise attention. Since the embedded labels are appended before the forward pass, the attention value towards the label corresponds to the attention map's last row excluding the label index. 

Furthermore, we average the attention maps across all samples, heads, and layers (similar to prior work by \citet{ye2025closerlooktabpfnv2}). We acknowledge that attention maps can vary substantially across these dimensions. However, this approach aligns with the intuition that features identified as relevant by the model across numerous samples, heads, or layers are those most indicative of importance (as we also show in our empirical results). In the following, the term ``attention score" of a feature refers to its average attention to the label column.

\section{Experiments and Results}
\label{experiment_results}

We now turn to the empirical analysis. 
First, we study TabPFN-Wide's performance in two settings: (a)~real-world HDLSS omics datasets (\autoref{subsec:real_world_results}) and (b)~standard benchmark tasks for predictive tabular machine learning (\autoref{subsec:benchmarks}) as well as a synthetic SNP dataset. Then, we assess its interpretability in \autoref{subsec:interpretability}.

\subsection{Datasets and Evaluation Protocol.}
We use machine learning–ready TCGA datasets differing from raw TCGA data by already being normalized, quality-checked, and otherwise pre-processed. We use five datasets: 
\textit{COAD}, \textit{LGG}, \textit{BRCA}, \textit{GBM} and \textit{OV} published by \citet{mlomicsbenchmark}.
Appendix A provides details of the corresponding table structures. Using early integration, we concatenate all omic types (mRNA, methylation, CNV (if present), and miRNA) along the feature axis, yielding datasets with up to $60{,}000$ features. In addition to these real-world datasets, we also evaluate on $21$ benchmark tasks (with $\leq10,000$ samples and $\leq500$ features) introduced by \textit{TabArena}~\citep{erickson2025tabarena}. We further extended our evaluation to 15 HDLSS datasets from \citep{li2018feature} as well as to synthetically generated single nucleotide polymorphism (SNP) datasets produced with HAPNEST \citep{HAPNEST}, which provide an HDLSS setting of up to 70,000 categorical features where the predictive signal is very sparse.



Unless stated otherwise, all models were evaluated using all features. If we apply feature reduction, we recursively merge features based on the minimal Euclidean distance of pairs of feature vectors (as demonstrated to be appropriate in preliminary analyses, see Appendix B).
We note that our main objective is to retain feature-wise interpretability and we solely explore it to compare model performance across different feature counts.

Alongside the foundation models TabPFNv2 and TabICL, we evaluate other baseline models, including the pre-tuned neural network RealMLP-TD \citep{holzmuller-realmlp-2025} as well as classical tree-based machine learning techniques like random forest and XGBoost~\citep{chen2016xgboost}. Importantly, ensembling was not used for TabPFN-Wide, TabPFNv2, TabICL, and RealMLP-TD to study raw model behaviour.

We perform 5-fold cross-validation for our biomedical datasets to compute AUROC and accuracy. 
For the TabArena datasets, we follow the original evaluation protocol and compute AUROC using a 3-fold cross-validation repeated 3 or 10 times, depending on the dataset size.

\subsection{Results on real-world wide datasets}
\label{subsec:real_world_results}


\begin{figure}[htbp]
    \centering
    \begin{minipage}[t]{0.42\linewidth}
        \vspace{0pt}
        \centering
        \includegraphics[width=\linewidth]{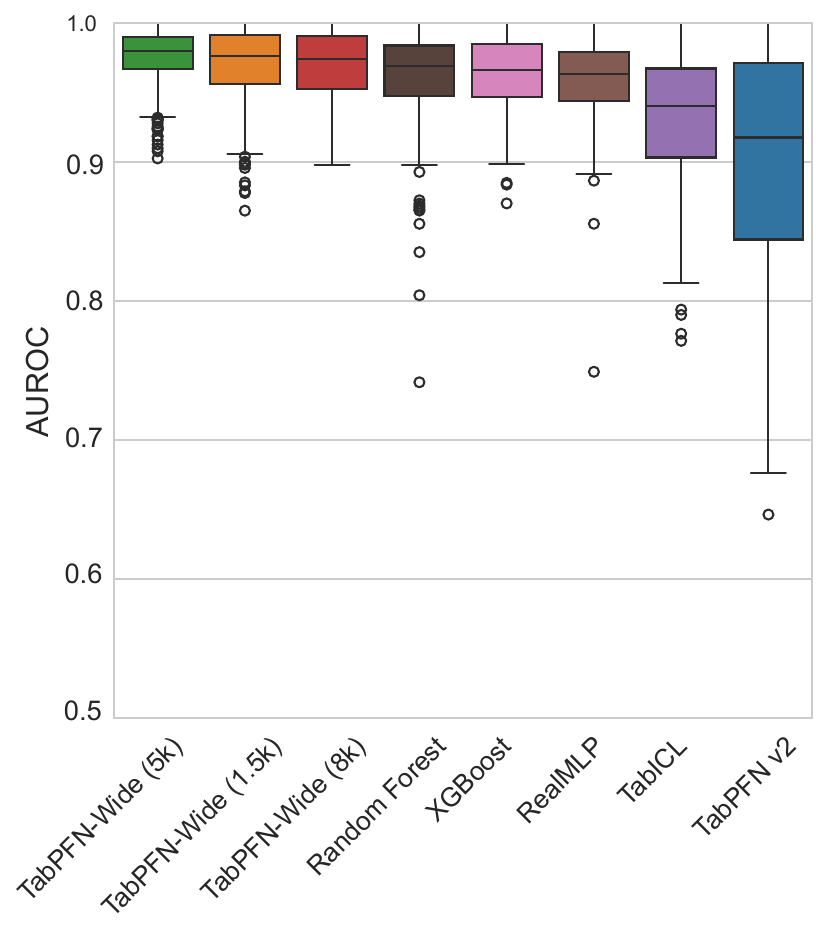}
        \caption{Average AUROC ($\pm$SD) scores on 5 multiomics cancer datasets.}
        \label{fig:relative_performance_multiomics}
    \end{minipage}
    \hfill
    \begin{minipage}[t]{0.54\linewidth}
        \vspace{0pt}
        \centering
        \includegraphics[width=\linewidth]{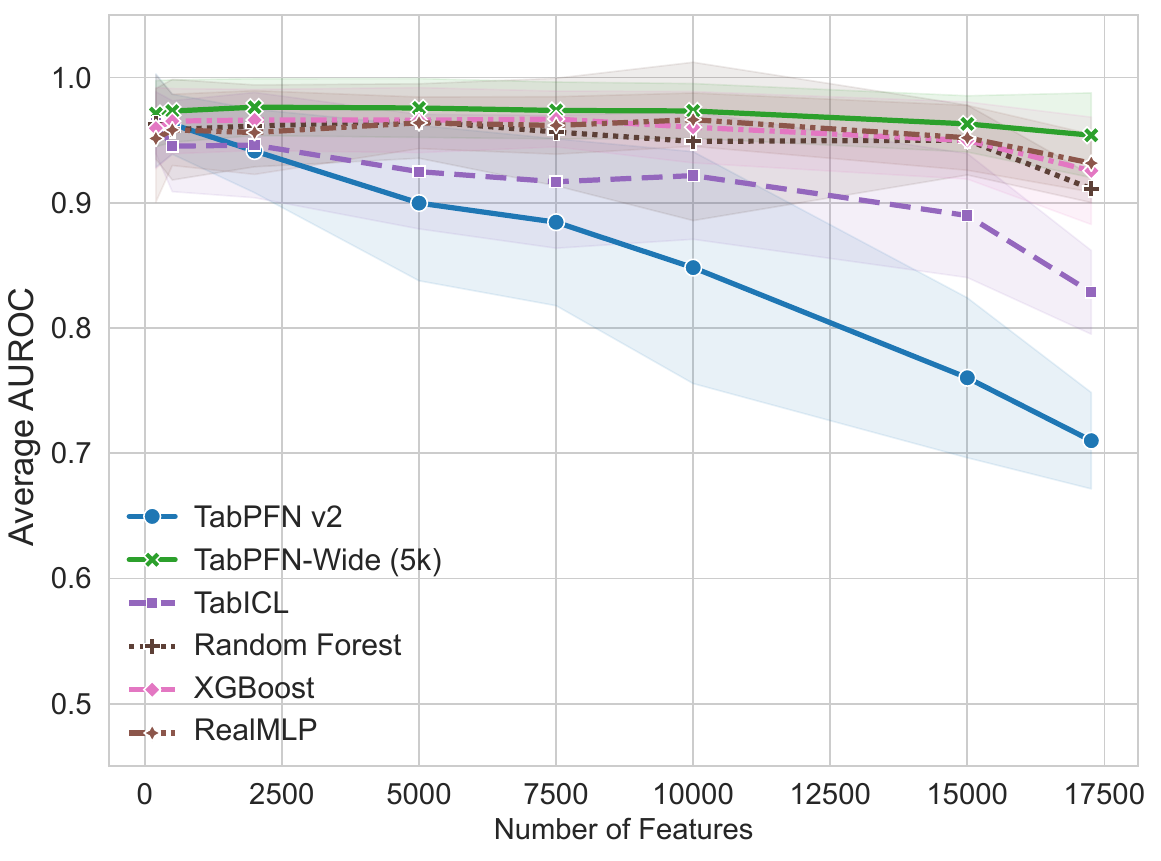}
        \caption{Average AUROC evaluated on five multiomics cancer datasets with feature reduction via agglomerative clustering.}
        \label{fig:omics_feature_reduction}
    \end{minipage}
\end{figure}
\noindent\textbf{TabPFN-Wide shows superior performance across real-world HDLSS datasets.} We first evaluated our models on the 5 TCGA cancer datasets from Yang \textit{et al.} on cancer subtype classification. The average AUROC scores in \cref{fig:relative_performance_multiomics} 
highlight the strong capabilities of TabPFN-Wide. While tree-based methods exhibit stable performance, our model achieves superior results. \todo{Update the plot to also show the tree based + realMLP}
TabPFNv2 and TabICL exhibit inferior performance consistent with the fact that they were not trained for such extreme feature counts. 

\noindent Interestingly, increasing the maximum width of synthetic datasets used during continued pre-training from $1{,}500$ to $8{,}000$ exerts only a minor influence on cancer subtype classification performance (\cref{fig:relative_performance_multiomics} and Appendix D), which is why we chose the 5k variant for all additional evaluations in the manuscript. Further evaluation is needed to assess the potential benefits of training on wider data, especially given the quadratic rise in complexity from increasing the number of features during training. 
Furthermore, we performed feature reduction to evaluate the performance trend of the models based on the number of available features. In this setting, TabPFN-Wide achieves the best overall results across increasing feature counts as seen in~\cref{fig:omics_feature_reduction}, remaining very stable while the performance of TabPFNv2 decreases significantly.

\subsection{Results on Standard Benchmarks and Widened Adaptations}
\label{subsec:benchmarks}

\begin{figure}[tbp]
    \centering
    
    \begin{minipage}[b]{0.42\linewidth}
        \centering
        \includegraphics[height=3.5cm]{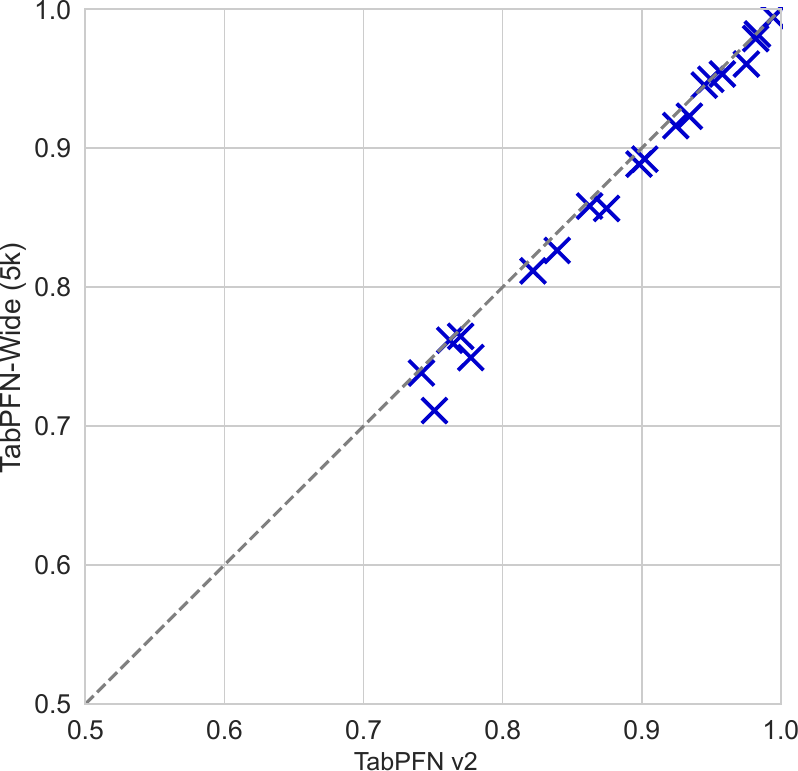}
        \\[1ex] \small (a)
        \label{fig:tabarena_vs_default}
    \end{minipage}
    \begin{minipage}[b]{0.48\linewidth}
        \centering
        \includegraphics[height=3.5cm]{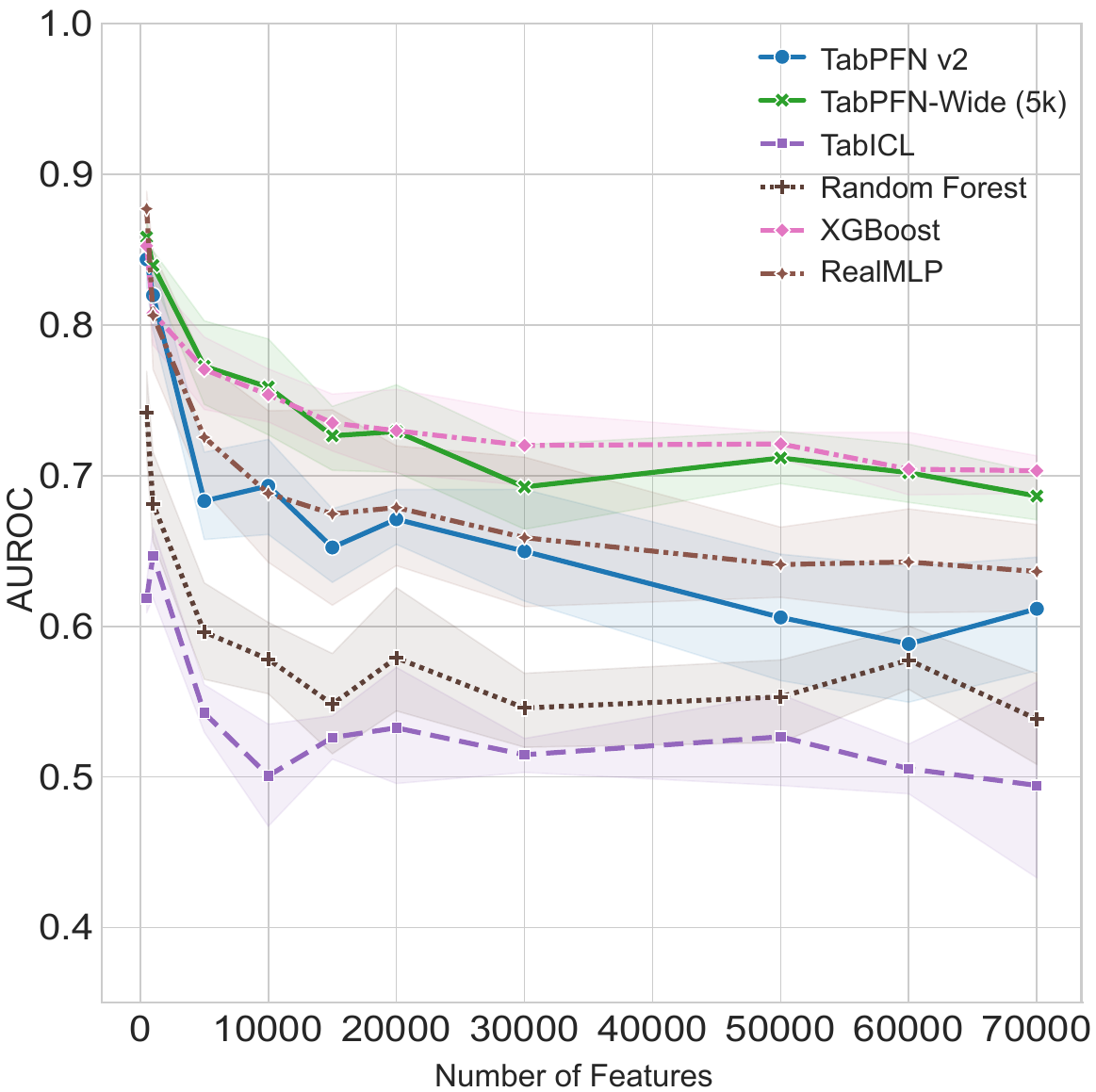}
        \\[1ex] \small (b)
        \label{fig:performance_snp}
    \end{minipage}
    
    \vspace{4ex} 
    \begin{minipage}[b]{\linewidth} 
        \centering
        \includegraphics[height=3.2cm]{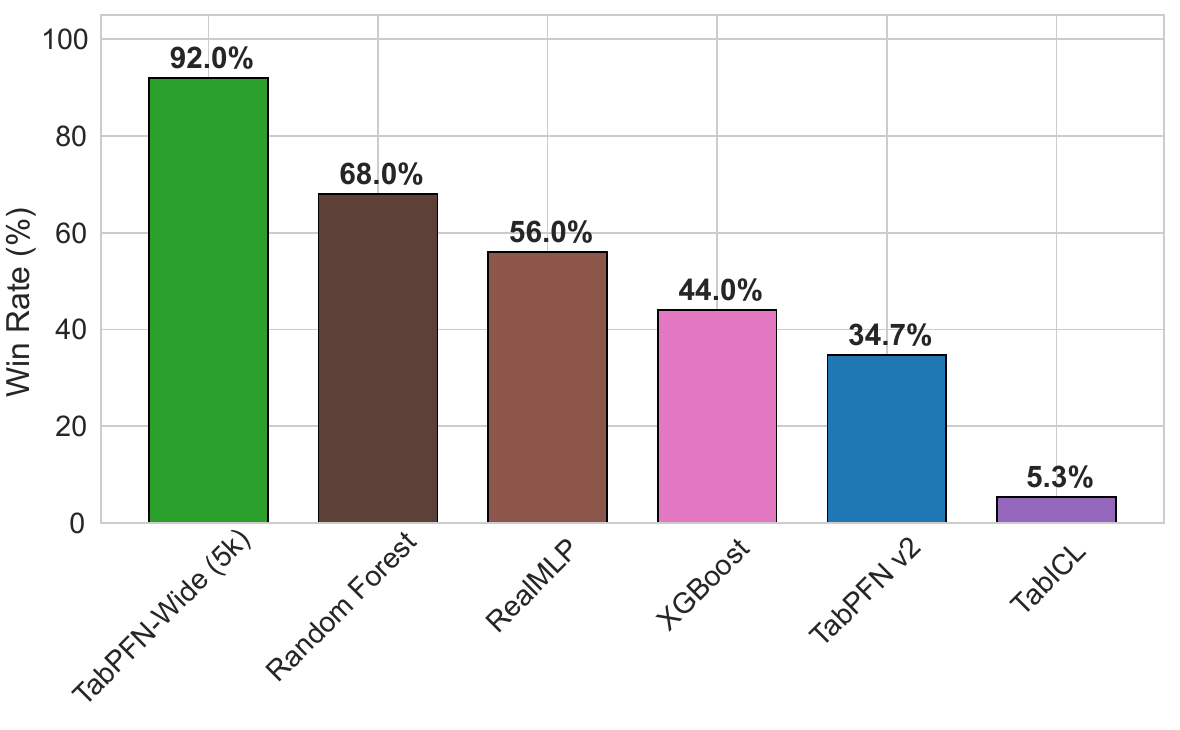}
        \\[1ex] \small (c)
        \label{fig:tabarena_hdlss}
    \end{minipage}
    
    \caption{(a) AUROC for TabPFN-Wide (5k) vs TabPFNv2 on $21$ TabArena classification tasks with \mbox{$\le10{,}000$} samples and \mbox{$\le500$} features. (b) Average AUROC for the SNP datasets with polygenicity of 0.01 (higher is better). We compare TabPFN-Wide, using up to 5k features for continued pre-training to TabPFNv2 and other baselines.(c) Aggregate AUROC-based pairwise win rate across 15 HDLSS datasets from \citep{li2018feature}, defined as the percentage of all pairwise comparisons where a given model achieved a strictly higher AUROC than its competitor.}
    \label{fig:tabarena}
\end{figure}

\textbf{TabPFN-Wide performs on par with TabPFNv2 on TabArena Benchmark.}  
\cref{fig:tabarena} (a) compares TabPFNv2 and TabPFN-Wide, showing that our continued pre-pretraining on wider datasets does not negatively impact performance on standard datasets with $\leq 10,000 $ samples and $\leq 500$ features (Spearman rho=0.9935). 
This suggests that there is no indication of catastrophic forgetting.

\noindent\textbf{Needle in a haystack.} We evaluate TabPFN-Wide on a biological noise-filtering task. Using SNP data, we generate binary phenotypes under a polygenic model
where only a low fraction (the \textit{polygenicity}) of SNPs are causal (See Appendix K for full details). To create a needle in a haystack scenario, we progressively increase the number of non-causal SNPs while keeping the set of causal variants fixed.

\Cref{fig:tabarena} b) reports the AUROC of the SNP datasets with polygenicity level of 0.01. As the number of non-causal SNPs increases, TabPFN-Wide and XGBoost exhibit the smallest degradation in performance. 
In contrast, TabICL is unable to reliably separate signal from noise, quickly converging toward random guessing ($\text{AUROC}=0.5$) as the feature dimensionality grows.


\noindent\textbf{HDLSS Benchmark from Li \textit{et al.} } Aggregate win rate analysis was used to evaluate relative model robustness across $15$ HDLSS datasets. 
As shown in \cref{fig:tabarena} (c), TabPFN-Wide achieves a $92.0\%$ win rate, substantially outperforming standard baselines (Random Forest, RealMLP, XGBoost) and Tabular Foundation Models (TabPFNv2, TabICL). This difference is also significant according to a paired Wilcoxon signed rank test comparing the AUCs ($p < 0.005$; See Appendix H for table of $p$-values). 

\subsection{Interpretability}
\label{subsec:interpretability}
%
To assess whether attention scores reflect feature importance, we used a controlled synthetic signal recovery benchmark with high-dimensional datasets containing a known subset of $k$ predictive features and compared attention-based rankings to impurity-based importances from a Random Forest. We report Recall@$k$—the proportion of truly predictive features among the top $k$ ranked—and analyze the mean importance gap between signal and noise features; across varying feature counts, numbers of informative features, and random seeds, TabPFN-Wide achieves Recall@$k$ and noise suppression on par with Random Forest (see Appendix F for details and plots).

Having evidence that attention maps yield useful insights in feature importance, we return to our real-world cancer datasets and validate the biological relevance of our model's attention scores by retrieving the features with the highest attention scores for subtype classification. Since mRNA is the most studied modality among the different omic types, we focus on the mRNA data. 
High correlation between genes complicates the task, since features that are presumably predictive are not necessarily causal. 

\textbf{TabPFN-Wide identifies important biomarkers for different cancer subtypes.} We extracted the $10$ genes with the highest attention scores from each dataset and examined their biological relevance according to literature (see Appendix A for details). In BRCA, nine genes show direct associations with breast cancer and one a general cancer link; in ovarian cancer, six are directly and two generally linked; for LGG and sarcoma, fewer direct associations (one and three, respectively) but more general cancer links were found, possibly reflecting limited prior study rather than lack of relevance, though variability in attention cannot be excluded. 
Overall, these results suggest that TabPFN-Wide’s attention scores capture meaningful feature importance signals and are able to recover biologically relevant biomarkers in cancer classification tasks.

\section{Conclusion}
\label{conclusion}
We introduce TabPFN-Wide, developed by continuing pre-training of TabPFNv2. To the best of our knowledge, it is the first tabular foundation model that handles HDLSS data without feature reduction and is the first application of continued pre-training to extend tabular foundation model capabilities. 
It achieves state-of-the-art performance on real-world and synthetic HDLSS data--demonstrating statistically significant improvements over standard baselines and existing foundation models--while simultaneously maintaining performance on small datasets. Furthermore, we show that attention scores, calculated within the transformer architecture, are indicative of feature importance and, thus, serve as an inherent interpretability method.

\subsection{Limitations and Outlook} Currently, our HDLSS prior is designed and validated only for continued pre-training of TabPFNv2. Initial attempts to train TabICL in the same manner were unsuccessful, raising the question of whether an adapted prior could solve this, or whether TabICL's architecture is inherently unable to handle HDLSS data (see Appendix I). Moreover, since the architecture of TabPFNv2 is unchanged, our model is limited by the (Flash-)attention mechanism's complexity and high memory requirements, restricting increases in the number of samples or features. Additionally, the attention map analysis may have limitations. Although this approach is highly accurate for synthetic problems where the ground truth is known (i.e., needle-in-a-haystack tasks), its applicability to realistic biomedical datasets should be interpreted with caution even though our results seem quite promising. 

Since our model is currently based solely on the TabPFNv2 classifier, our approach seeks further validation from continuing pre-training of the regressor model. The prior setup is strongly inspired by the type of data faced in the biomedical domain, raising questions about whether a more advanced HDLSS prior allows the creation of an even better TabPFN-Wide. While our findings suggest that attention scores are a valid approach for inherent interpretability, a systematic evaluation will be future work. Overall, we show that continued pre-training has the potential to extend the capabilities of pre-trained models, like TabPFNv2, paving the way for resource-efficient generation of ``patched" model versions for other dataset characteristics and that TabPFN-Wide is a promising method for many future studies with tabular data, such as in biomedicine. 


\begin{appendices}

\end{appendices}

\section{Competing interests}
No competing interest is declared.

\section{Author contributions statement}
C.K., K.E., and N.P. conceived the experiment(s),  C.K., J.K., J.H, and S.O. conducted the experiment(s) and analysed the results. K.E. and N.P. supervised the experiments and provided additional ideas. All authors wrote and reviewed the manuscript.

\section{Acknowledgments}
This work was supported by the Deutsche Forschungsgemeinschaft (DFG, German Research Foundation) under Germany's Excellence Strategy—EXC number 2064/1—Project number 390727645.

\bibliography{reference_condensed}

\bibliographystyle{abbrvnat}

\clearpage
\onecolumn
\begin{bibunit}[abbrvnat]
\section*{Appendix A: Data Overview Multiomics Datasets}
\subsection*{Data Overview}
\label{app:data_overview}
\cref{tab:data_overview} gives an overview of the number of samples and features of the omics datasets. Furthermore, it shows which molecular measurements are available for which dataset. Datasets provided by \citet{mlomicsbenchmark} (LGG, OV, COAD) have 4 different omics: mRNA gene expression data (mRNA), copy number variation data (CNV), methylation data (Methylation) and micro RNA data (miRNA). MRNA, CNV, and methylation features are measurements corresponding to human genes. For our usage, we concatenated all different omics resulting in up to $60{,}000$ features. Datasets provided by \citet{shamirdata} consist of less features due to missing CNV data and lower number of features for methylation data.

\begin{table}[H]
\centering
\begin{tabular}{l|llllll}
\toprule
        & Patients & mRNA  & CNV   & Methylation & miRNA & All   \\ 
\midrule
LGG \scriptsize{(low grade glioma)}    & 247      & $14{,}260$ & $21{,}104$ & $24{,}979$       & 321   & $60{,}664$ \\
OV \scriptsize{(ovarian cancer)}     & 284      & $14{,}229$ & $21{,}104$ & $24{,}797$       & 313   & $60{,}443$ \\
COAD \scriptsize{(colon adetrcinoma)}     & 260      & $17{,}261$ & $19{,}551$ & $19{,}052$       & 375   & $56{,}239$ \\
\midrule
BRCA \scriptsize{(breast cancer)} & 440      & $20{,}531$ & N/A   & $5{,}000$        & $1{,}046$  & $26{,}577$ \\
SARC \scriptsize{(sarcoma)} & 259      & $20{,}531$ & N/A   & $5{,}000$        & $1{,}046$  & $26{,}577$ \\
GBM \scriptsize{(glioblastoma)}     & 274      & $12{,}042$ & N/A   & $5{,}000$        & 534   & $17{,}576$ \\
\bottomrule

\end{tabular}
\caption{Number of samples and features for all used datasets. SARC is only used for attention analysis.}
\label{tab:data_overview}
\end{table}
\newpage
\subsection*{Genes with highest attention scores}

\label{subsec:other_datasets_important_genes}

As described in the interpretability section, we analyzed the genes with the highest attention scores from our datasets with respect to literature connecting the gene with the given cancer type. We classified each gene as (i) directly associated with the specified cancer subtype, (ii) generally associated with cancer across multiple types, or (iii) having no known association with cancer. As this analysis was conducted manually, the list of citations should not be considered exhaustive. In cases where a PubMed search did not yield relevant literature, no potential associations were reported.

\begin{table}[H]

    \centering

    \label{tab:important_genes_app}

    \begin{tabular}{p{0.1\linewidth}p{0.25\linewidth}p{0.25\linewidth}>{\raggedright\arraybackslash}p{0.2\linewidth}}

        \toprule

        \textbf{Dataset} & 
        
        \textbf{Direct Connection} & 

        \textbf{General Connection to Cancer} & 

        \textbf{No Known Connection} \\

        \midrule
        BRCA & FOXC1~\citep{FOXC1},\newline  FOXA1~\citep{RN5153},\newline SFT2D2~\citep{RN5154},\newline ESR1~\citep{RN5155},\newline CENPA~\citep{RN5156},\newline FAM171A1~\citep{RN5157},\newline TPX2~\citep{RN5158}, \newline CCDC170~\citep{RN5159}, \newline GATA3~\citep{RN5160} & SRSF12~\cite{RN5152} & \\

        \bottomrule
        LGG & NAPE-PLD~\citep{wu2012alteration} &
        MIR1307~\citep{sumer2025selective} \newline
        CCDC177~\citep{kumar2018methylation}~\citep{ju2020genome} \newline
        MET~\citep{cheng2019met}, \newline
        MIER1~\citep{clements2012differential}, \newline
        GPN1~\citep{zhu2024comprehensive}
        & LOC101928075,\newline C4B, \newline ZZZ696, \newline PRKAR1B  \\

        \bottomrule

        OV & 
        CMPK1~\citep{zhou2017cytidine}, \newline
        PLEKHA5~\citep{singh2018genome}, \newline
        LOC101927151~\citep{zheng2020identification}, \newline
        GATA6.AS1~\citep{xu2021gata6}, \newline
        MT1F~\citep{murakami2008mta1}, \newline
        ETFDH~\citep{wang2023identification}, \newline

        &
        PAFAH1B1~\citep{lo2012overexpression}\citep{majmudar2025neural}, \newline
        RAB24~\citep{ding2025ras}
        &
        CCDC40, \newline
        LOC101928069 \\

        \bottomrule
        SARC &
        COL22A1~\citep{pan2022novel} \newline
        GNPNAT1~\citep{tolwani2021prognostic} \newline
        ARHGAP42~\citep{dermawan2023malignant} \newline &
        TPCN2~\citep{alharbi2019endolysosomal} \newline
        DPEP3~\citep{hamilton2020tamrintamab} \newline
        MRPL46~\citep{wu2025mitochondrial} \newline
        TAS2R19~\citep{carey2022t2r} \newline
        TCEB3~\citep{cai2024tceb3} \newline
        MON1B~\citep{jiang2018knockdown} \newline
        FGFR1OP2~\citep{yang2022fgfr1op2}

    \end{tabular}

    \caption{Categorization of the top 10 features with the highest attention scores for datasets when performing subtype classification.}
\end{table}
\section*{Appendix B: Comparison of different feature reduction techniques}
\begin{figure}[H]
    \centering
    
    \begin{minipage}[b]{0.45\textwidth}
        \centering
        \includegraphics[width=\textwidth]{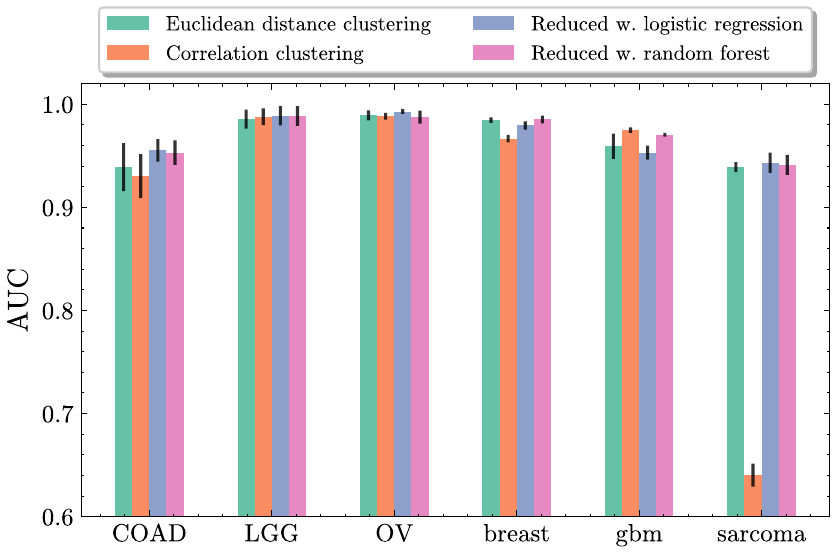}
        \\ \small (a)
        \label{fig:500_features}
    \end{minipage}
    \hfill
    \begin{minipage}[b]{0.45\textwidth}
        \centering
        \includegraphics[width=\textwidth]{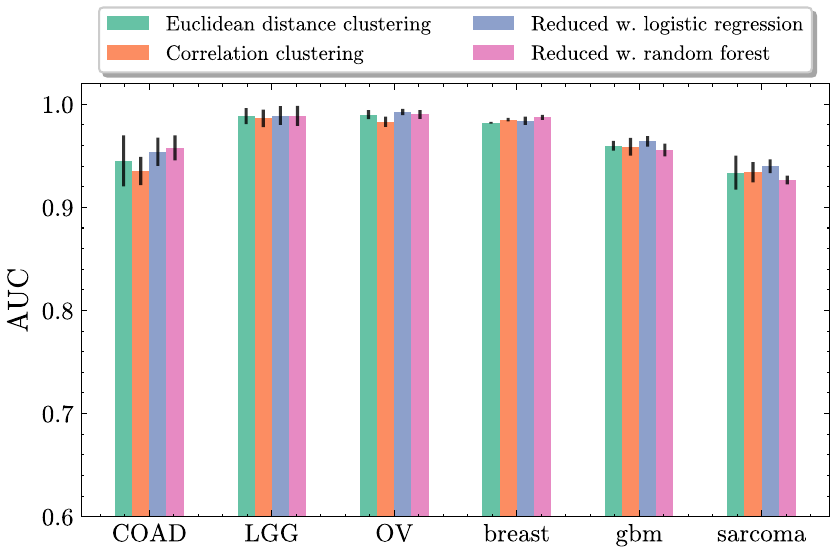}
        \\ \small (b)
        \label{fig:2000_features}
    \end{minipage}
    
    \caption{AUROC of TabPFNv2 evaluated on different datasets reduced to (a)~$500$ features and (b)~$2{,}000$ features using different techniques.}
    \label{fig:feature_reduction}
\end{figure}

In preliminary experiments, we tested the performance of TabPFNv2 on our real-world HDLSS datasets reduced with different feature reduction methods. Since this is not our main priority, we focused on simple approaches offered by \textit{sci-kit learn}. Although we tested both supervised (label-based) and unsupervised feature reduction methods, our preference was for the unsupervised approaches, as they better mitigate the risk of overfitting in HDLSS settings. For biomedical data, a common approach is to cluster by correlation which we compared against clustering by lowest Euclidean distance between feature vectors and reduction using the feature importance weights from fitted machine learning models. Given that Euclidean distance-based clustering frequently outperforms the correlation-based approach for our data (see \cref{fig:feature_reduction}) and achieves performance comparable to supervised methods, we adopted this strategy for our analyses.


    
    

\section*{Appendix C: Detailed results for all multiomics datasets}

\begin{figure}[H]
    \centering
    \begin{minipage}{0.45\textwidth}
        \centering
        \includegraphics[width=\linewidth]{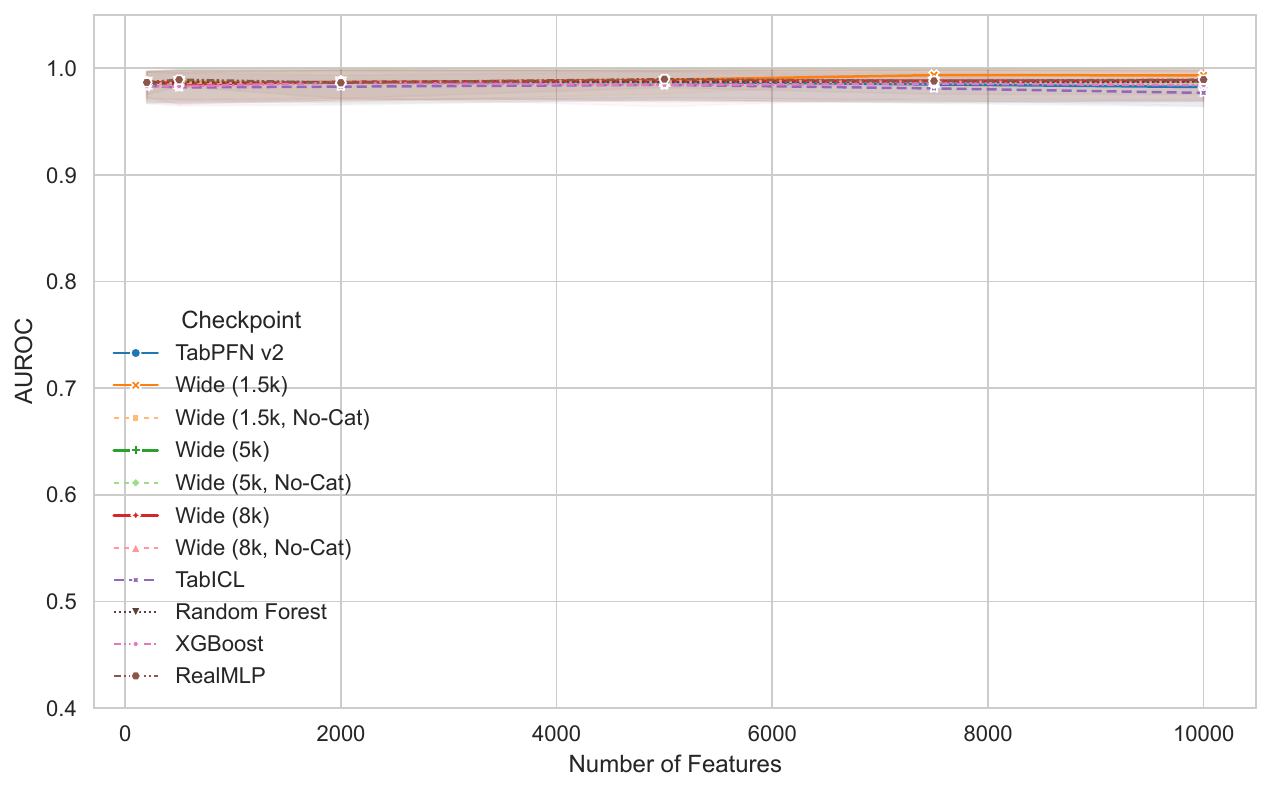}
        \\ \small (a) LGG
    \end{minipage}
    \hfill
    \begin{minipage}{0.45\textwidth}
        \centering
        \includegraphics[width=\linewidth]{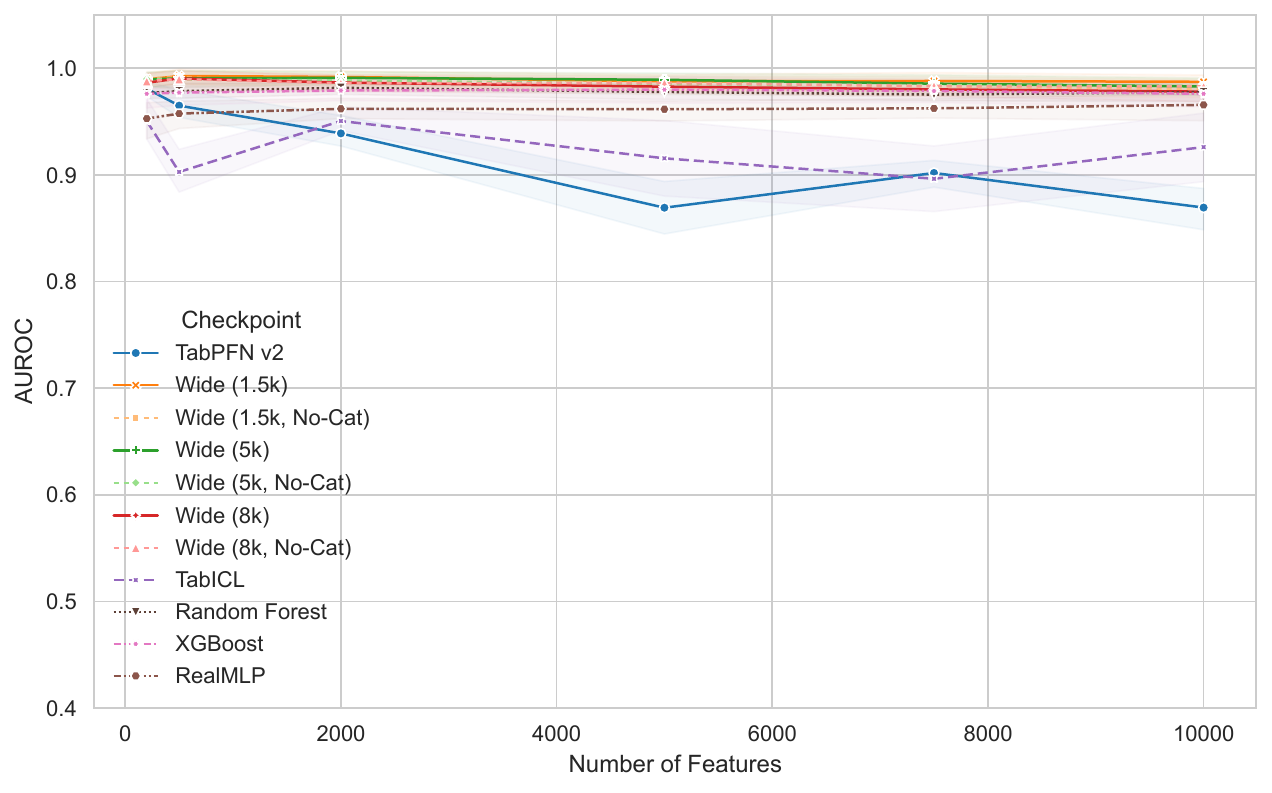}
        \\ \small (b) OV
    \end{minipage}

    \begin{minipage}{0.45\textwidth}
        \centering
        \includegraphics[width=\linewidth]{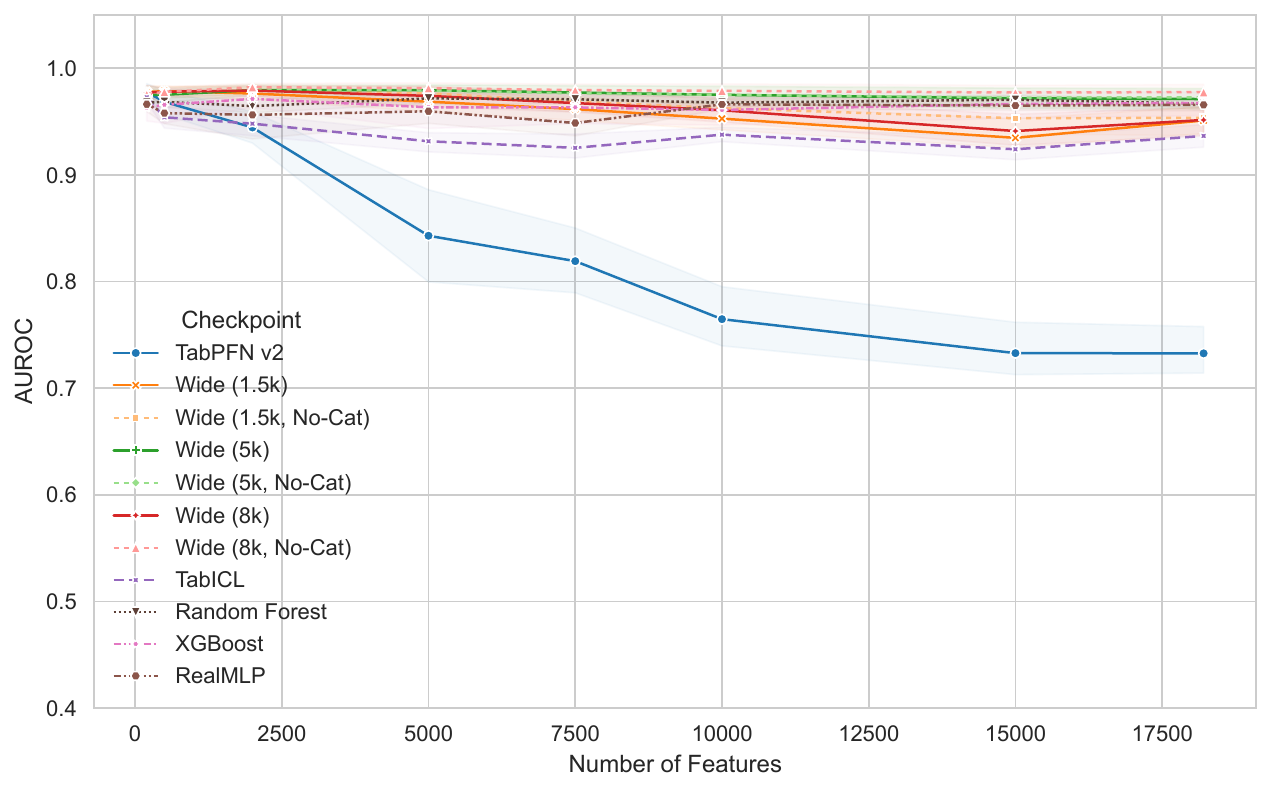}
        \\ \small (c) BRCA
    \end{minipage}
    \hfill
    \begin{minipage}{0.45\textwidth}
         \centering
         \includegraphics[width=\linewidth]{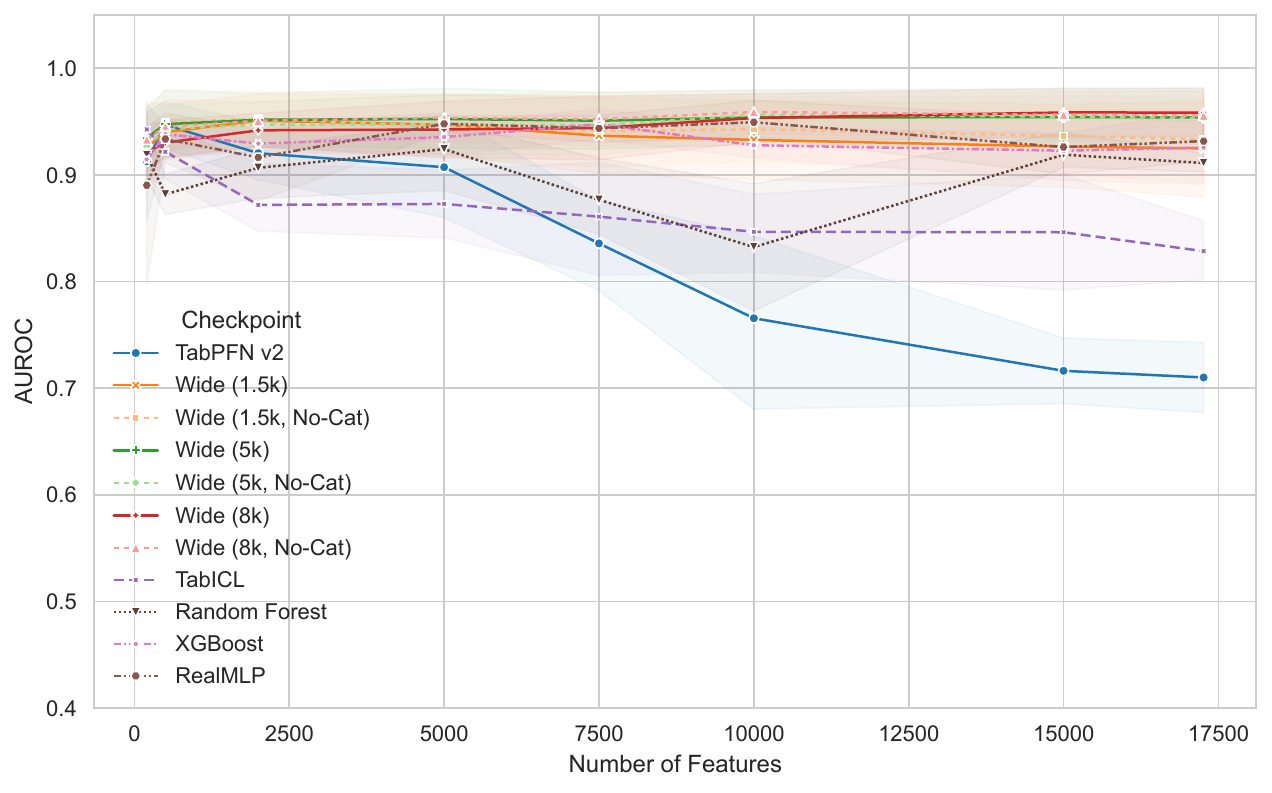}
        \\ \small (d) COAD
    \end{minipage}

    \caption{Single results for all datasets with feature reduction applied. The axis were chosen such that the differences in feature numbers and AUROC scores becomes comparable.}
    \label{fig:multiomics_feature_reduction_detailed}
\end{figure}
\newpage
\section*{Appendix D: Multiomics}
\label{app:different_metrics}
\begin{table}[h]
\centering

\begin{tabular}{lccccc}
\toprule
Model & BRCA & COAD & GBM & LGG & OV \\
\#features & 18,206 & 17,261 & 18,614 & 14,260 & 14,229 \\
\midrule
TabPFN v2 & $0.733 \pm 0.029$ & $0.710 \pm 0.038$ & $0.794 \pm 0.054$ & $0.973 \pm 0.018$ & $0.836 \pm 0.036$ \\
Wide (1.5k) & $0.951 \pm 0.017$ & $0.925 \pm 0.051$ & $0.941 \pm 0.017$ & $0.988 \pm 0.009$ & $\bm{0.987 \pm 0.014}$ \\
Wide (1.5k, No-Cat) & $0.954 \pm 0.018$ & $0.933 \pm 0.047$ & $0.960 \pm 0.018$ & $0.986 \pm 0.022$ & $0.977 \pm 0.013$ \\
Wide (5k) & $0.971 \pm 0.012$ & $0.954 \pm 0.034$ & $0.962 \pm 0.021$ & $0.988 \pm 0.022$ & $0.981 \pm 0.010$ \\
Wide (5k, No-Cat) & $0.973 \pm 0.012$ & $0.953 \pm 0.032$ & $0.961 \pm 0.019$ & $0.988 \pm 0.020$ & $0.982 \pm 0.012$ \\
Wide (8k) & $0.952 \pm 0.017$ & $\bm{0.959 \pm 0.027}$ & $0.936 \pm 0.021$ & $0.987 \pm 0.021$ & $0.982 \pm 0.011$ \\
Wide (8k, No-Cat) & $\bm{0.978 \pm 0.009}$ & $0.956 \pm 0.030$ & $\bm{0.965 \pm 0.024}$ & $0.987 \pm 0.021$ & $0.982 \pm 0.010$ \\
\midrule
TabICL & $0.937 \pm 0.013$ & $0.828 \pm 0.033$ & $0.898 \pm 0.020$ & $0.973 \pm 0.021$ & $0.883 \pm 0.033$ \\
Random Forest & $0.967 \pm 0.005$ & $0.911 \pm 0.011$ & $0.954 \pm 0.025$ & $0.986 \pm 0.017$ & $0.974 \pm 0.011$ \\
XGBoost & $0.967 \pm 0.015$ & $0.926 \pm 0.043$ & $0.951 \pm 0.042$ & $0.987 \pm 0.016$ & $0.969 \pm 0.020$ \\
RealMLP & $0.966 \pm 0.007$ & $0.932 \pm 0.023$ & $0.960 \pm 0.028$ & $\bm{0.991 \pm 0.008}$ & $0.970 \pm 0.018$ \\
\bottomrule
\end{tabular}

\caption{Average AUROC ($\pm$SEM) scores of 5 multiomics datasets (higher is better). We compare TabPFN-Wide, using up to 8k features for continued pre-training, to TabPFNv2 and other baseline methods and boldface the best values for each column.}
\label{tab:average_auroc_multiomics}
\end{table}
\begin{table}[h]
\centering

\begin{tabular}{lccccc}
\toprule
Model & BRCA & COAD & GBM & LGG & OV \\
\#features & 18,206 & 17,261 & 18,614 & 14,260 & 14,229 \\
\midrule
TabPFN v2 & $0.528 \pm 0.006$ & $0.669 \pm 0.009$ & $0.303 \pm 0.010$ & $0.887 \pm 0.048$ & $0.514 \pm 0.049$ \\
Wide (1.5k) & $0.744 \pm 0.022$ & $0.835 \pm 0.017$ & $0.717 \pm 0.050$ & $0.919 \pm 0.020$ & $0.877 \pm 0.075$ \\
Wide (1.5k, No-Cat) & $0.797 \pm 0.025$ & $0.862 \pm 0.016$ & $0.771 \pm 0.070$ & $\bm{0.976 \pm 0.026}$ & $0.852 \pm 0.027$ \\
Wide (5k) & $0.852 \pm 0.032$ & $0.858 \pm 0.017$ & $0.799 \pm 0.071$ & $0.964 \pm 0.009$ & $0.880 \pm 0.049$ \\
Wide (5k, No-Cat) & $0.852 \pm 0.028$ & $0.873 \pm 0.011$ & $0.787 \pm 0.088$ & $0.960 \pm 0.000$ & $0.877 \pm 0.050$ \\
Wide (8k) & $0.759 \pm 0.017$ & $0.869 \pm 0.042$ & $0.737 \pm 0.059$ & $0.972 \pm 0.011$ & $0.866 \pm 0.064$ \\
Wide (8k, No-Cat) & $0.864 \pm 0.029$ & $0.873 \pm 0.017$ & $\bm{0.816 \pm 0.063}$ & $0.964 \pm 0.009$ & $\bm{0.887 \pm 0.034}$ \\
\midrule
TabICL & $0.767 \pm 0.056$ & $0.781 \pm 0.073$ & $0.660 \pm 0.045$ & $0.911 \pm 0.037$ & $0.676 \pm 0.063$ \\
Random Forest & $0.815 \pm 0.014$ & $0.831 \pm 0.042$ & $0.779 \pm 0.092$ & $0.960 \pm 0.014$ & $0.859 \pm 0.047$ \\
XGBoost & $\bm{0.867 \pm 0.026}$ & $0.862 \pm 0.037$ & $0.775 \pm 0.087$ & $0.964 \pm 0.017$ & $0.838 \pm 0.072$ \\
RealMLP & $0.845 \pm 0.016$ & $\bm{0.892 \pm 0.022}$ & $0.783 \pm 0.094$ & $0.955 \pm 0.023$ & $0.827 \pm 0.041$ \\
\bottomrule
\end{tabular}

\caption{Average accuracy ($\pm$SEM) scores of 5 multiomics datasets (higher is better). We compare TabPFN-Wide, using up to 8k features for continued pre-training (second column), to TabPFNv2 and other baseline methods and boldface the best values for each column.}
\label{tab:multiomics_accuracy}
\end{table}
\newpage

\section*{Appendix E: Unlearning Analysis}
\begin{figure}[H]
   \centering
    
    \begin{minipage}{0.35\textwidth}
        \centering
      \includegraphics[width=\linewidth]{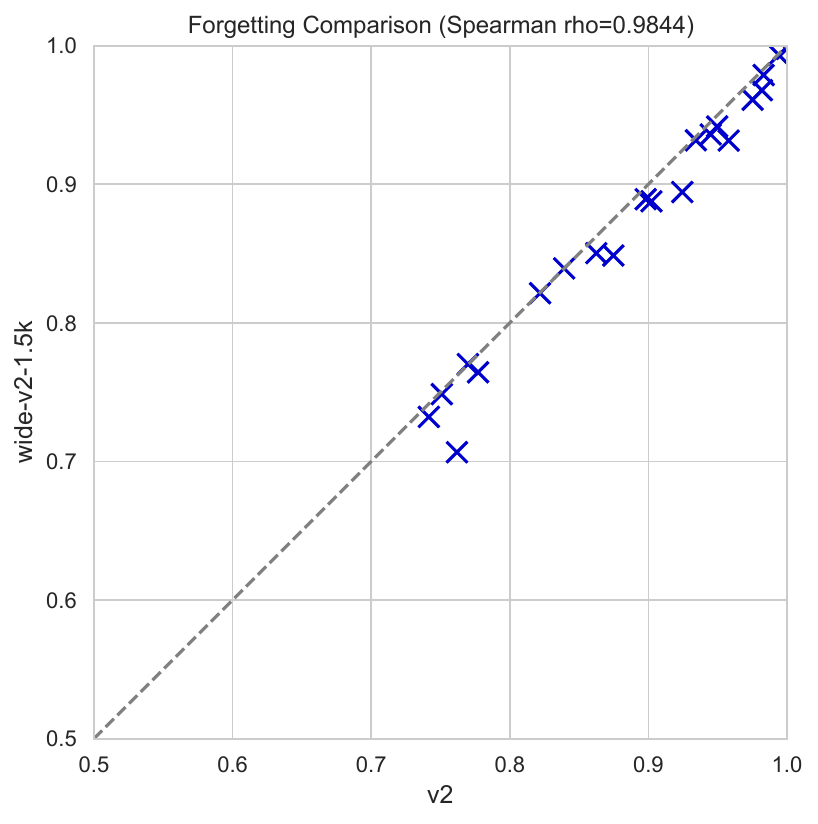}
        \\ \small (a) Unlearning Scatter plot TabPFNv2 vs TabPFN Wide 1.5k
        \label{fig:unlearning_scatter_1_5k}
    \end{minipage}
    \begin{minipage}{0.35\textwidth}
        \centering
      \includegraphics[width=\linewidth]{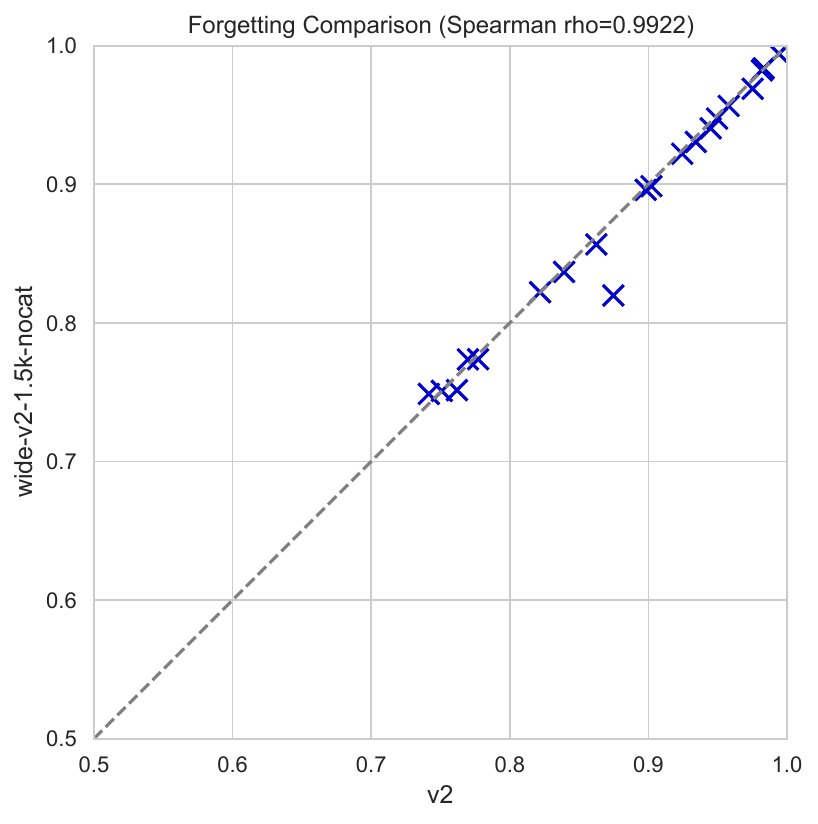}
        \\ \small (b) Unlearning Scatter TabPFNv2 vs TabPFN Wide 1.5k non categorical
        \label{fig:unlearning_scatter_1_5k_nocat}
    \end{minipage}
     \begin{minipage}{0.35\textwidth}
        \centering
      \includegraphics[width=\linewidth]{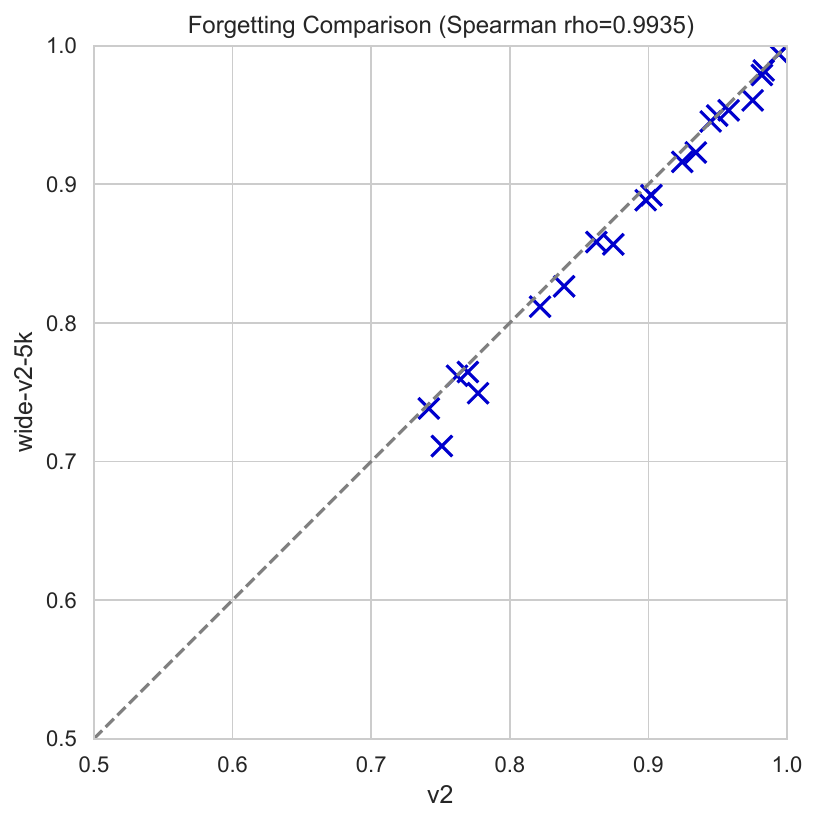}
        \\ \small (c) Unlearning Scatter TabPFNv2 vs TabPFN Wide 5k
    \end{minipage}
    \begin{minipage}{0.35\textwidth}
        \centering
      \includegraphics[width=\linewidth]{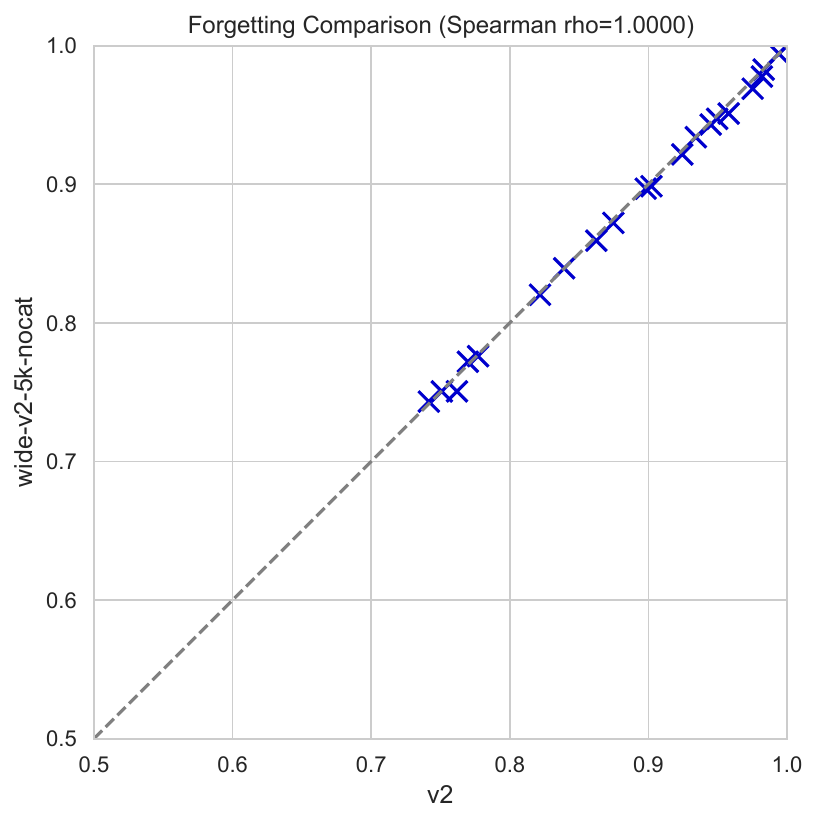}
        \\ \small (d) Unlearning Scatter TabPFNv2 vs TabPFN Wide 5k non categorical
    \end{minipage}
     \begin{minipage}{0.35\textwidth}
        \centering
      \includegraphics[width=\linewidth]{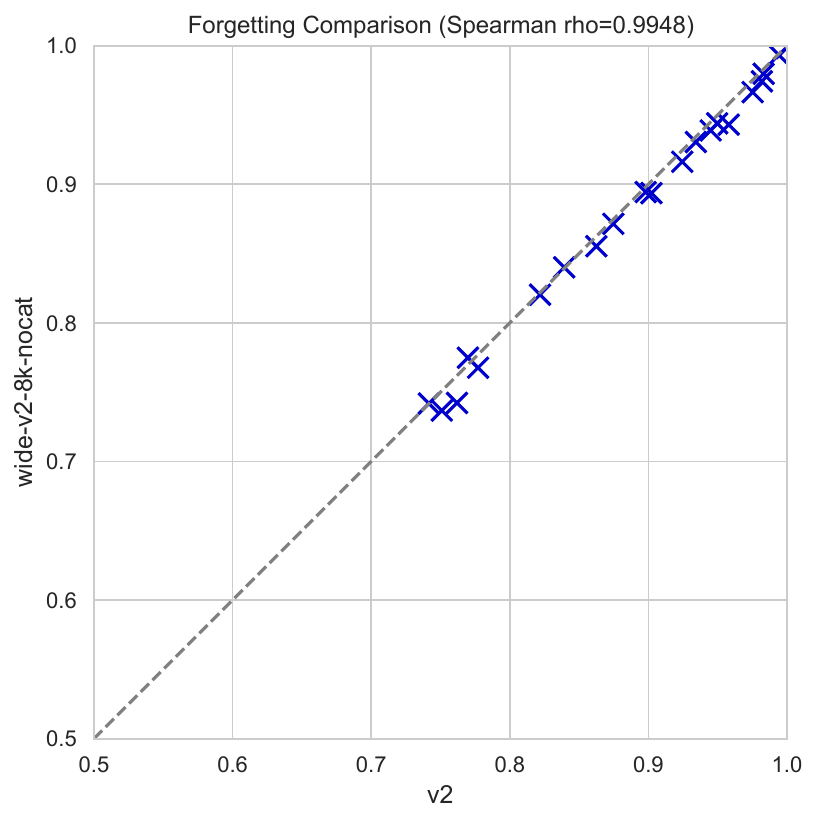}
        \\ \small (e) Unlearning Scatter TabPFNv2 vs TabPFN Wide 8k
    \end{minipage}
    \begin{minipage}{0.35\textwidth}
        \centering
      \includegraphics[width=\linewidth]{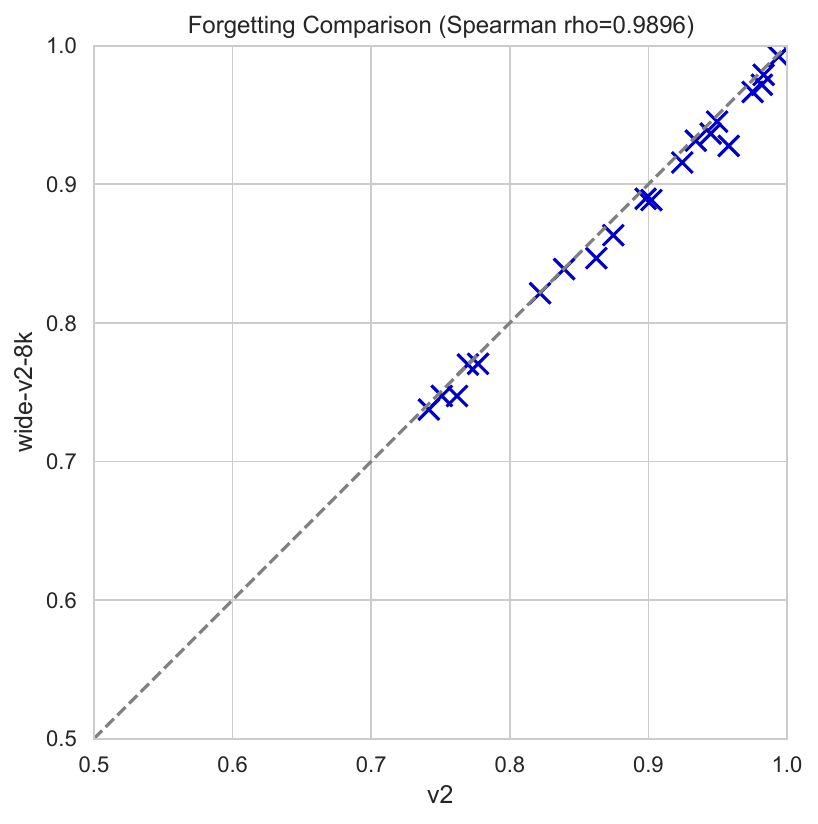}
        \\ \small (f) Unlearning Scatter TabPFNv2 vs TabPFN Wide 8k non categorical
    \end{minipage}
    \caption{}
\end{figure}

\section*{Appendix F: Synthetic Benchmark for Feature Importance}

    
To quantitatively evaluate whether the attention scores produced by TabPFN-Wide reliably capture feature importance, we conducted a controlled synthetic benchmark. The goal of this experiment is to test the model's ability to isolate a small number of true predictive features (signal) from a large pool of uninformative features within a low sample size regime. 

We generated binary classification datasets using \texttt{scikit-learn}'s \texttt{make\_classification} function. To simulate challenging scenarios, we fixed the number of samples to 50 and varied the total number of features across $\{200, 500, 1000, 3000, 5000\}$. To test different levels of signal sparsity, the number of truly informative features, $k$, was varied across $\{3, 5, 7\}$. The classes were generated with a class separation factor of 1.5, and no redundant or repeated features were included. 

For each combination of feature count and informative features, we generated datasets across 5 independent random seeds to ensure statistical robustness. We evaluated TabPFN-Wide (using the 5k model) against a Random Forest baseline configured with 200 estimators. 

\begin{figure}[H]
    \centering
    
    \begin{minipage}[b]{0.4\textwidth}
        \centering
        \includegraphics[width=\textwidth]{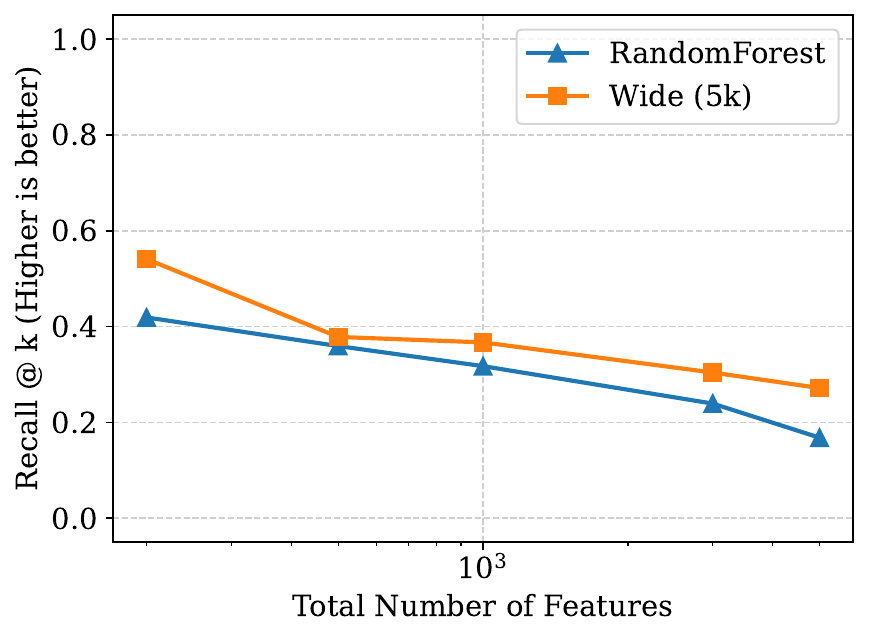}
        \\ \small (a)
        \label{fig:feature_importance_recall_at_k}    
    \end{minipage}
    \hfill
    \begin{minipage}[b]{0.4\textwidth}
        \centering
        \includegraphics[width=\textwidth]{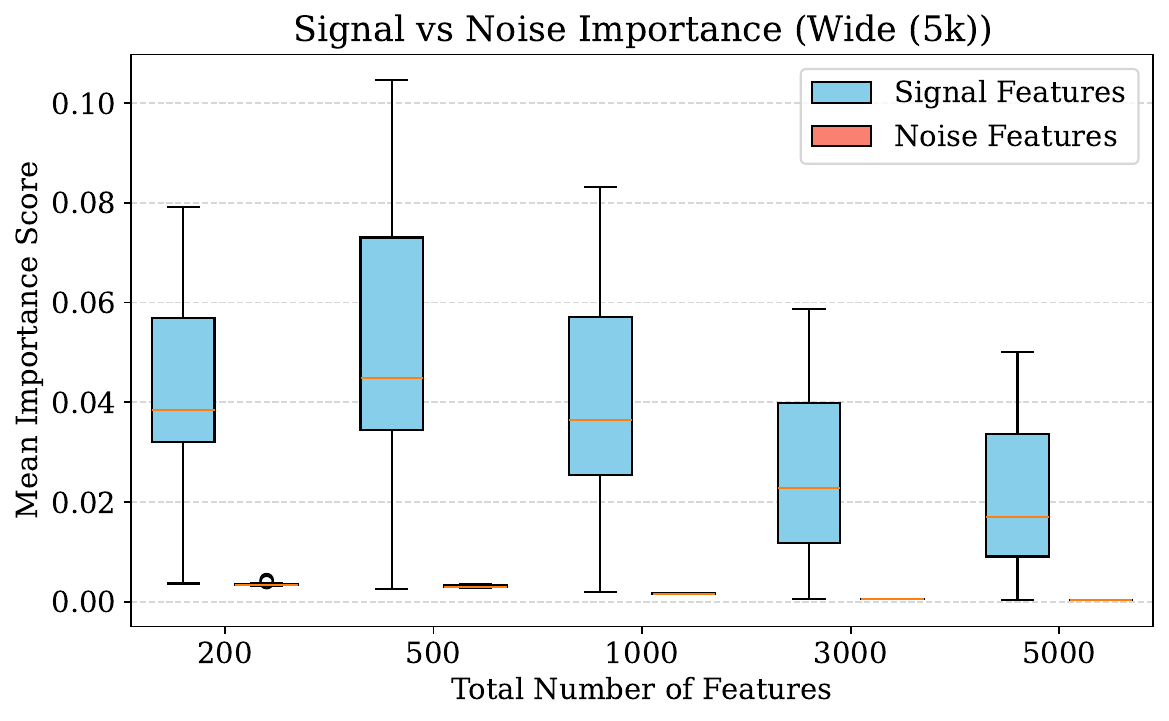}
        \\ \small (b)
        \label{fig:boxplot comparison signal noise TabPFN Wide}
    \end{minipage}
    \caption{Recall@$k$ for TabPFN-Wide and Random Forest across varying feature dimensions (a) and distribution of mean importance scores for signal versus noise features using TabPFN-Wide (b).}
    \label{fig:attention_comparison_noise}
\end{figure}

We assessed the feature ranking capabilities of both models using two primary approaches:
\begin{itemize}
    \item \textbf{Recall@$k$}: We extracted the top $k$ features ranked by attention score (for TabPFN-Wide) or impurity (for Random Forest) and calculated the proportion of true informative features successfully recovered within this top subset (see \cref{fig:attention_comparison_noise} (a)). Because $k$ exactly matches the number of true informative features in each setting, this serves as a strict recovery metric.
    \item \textbf{Signal-to-Noise Separation}: We computed the mean importance score assigned to the true signal features versus the uninformative noise features to visualize the noise floor (see \cref{fig:attention_comparison_noise} (b)).
\end{itemize}

\section*{Appendix G: Overview Results TabArena}
\begin{figure}[H]
   \centering
    
    \begin{minipage}{\textwidth}
        \centering
      \includegraphics[width=\linewidth]{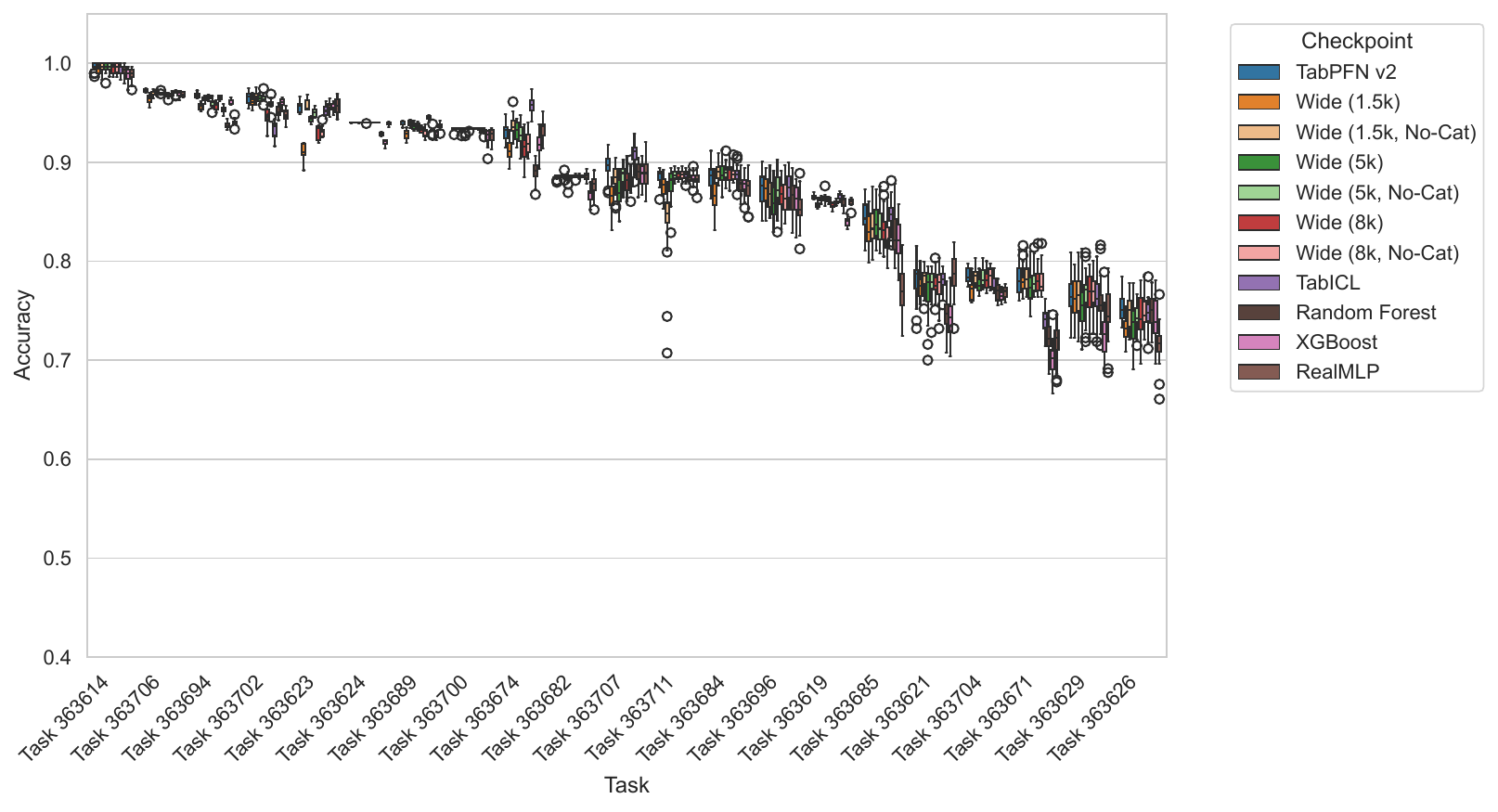}
        \\ \small (a) Accuracy per Task
    \end{minipage}
    \hfill
    \begin{minipage}{\textwidth}
        \centering
      \includegraphics[width=\linewidth]{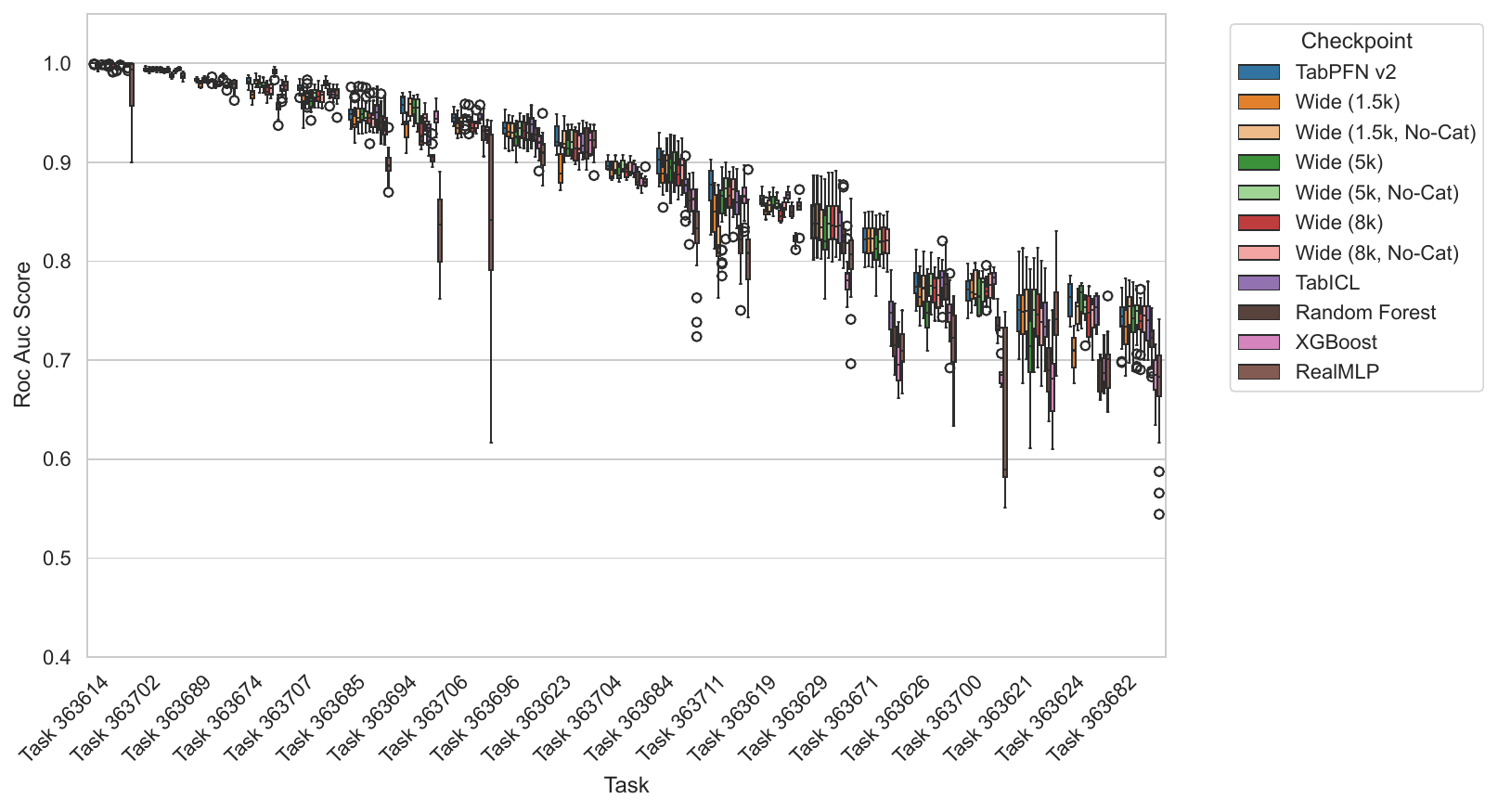}
        \\ \small (b) ROC-AUC per Task
        \label{fig:tabarena_detailed_metrics}
    \end{minipage}
    \hfill

    \caption{Box plots visualizing the comparison of TabPFNWide in its different variants and other models in terms of Accuracy and ROC-AUC.}
    \label{fig:openml_additional}
\end{figure}

\section*{Appendix H: Overview Results on HDLSS data from \citep{li2018feature}}
\begin{figure}[H]
   \centering
    
    \begin{minipage}{\textwidth}
        \centering
      \includegraphics[width=\linewidth]{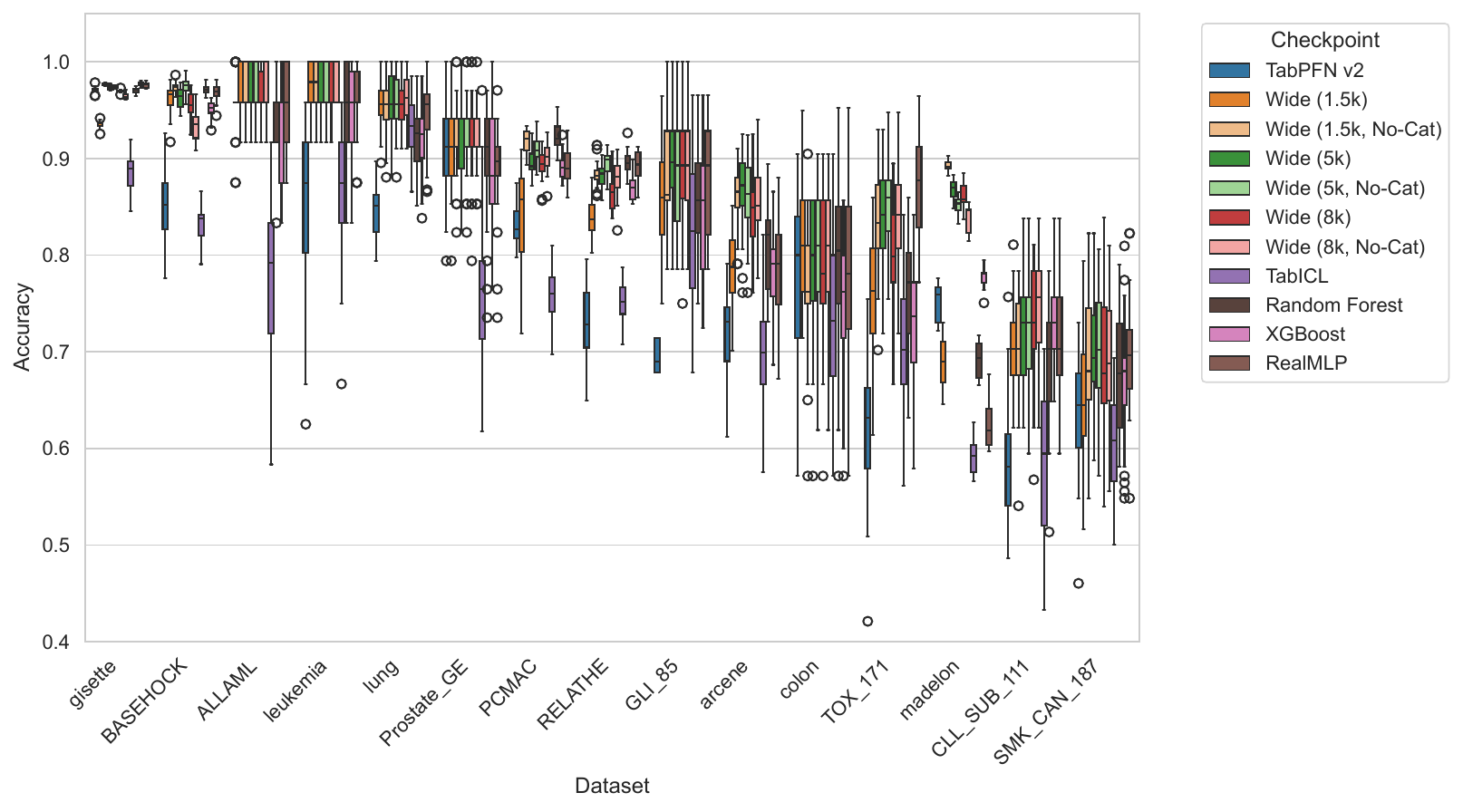}
        \\ \small (a) Accuracy per Dataset
    \end{minipage}
    \hfill
    \begin{minipage}{\textwidth}
        \centering
      \includegraphics[width=\linewidth]{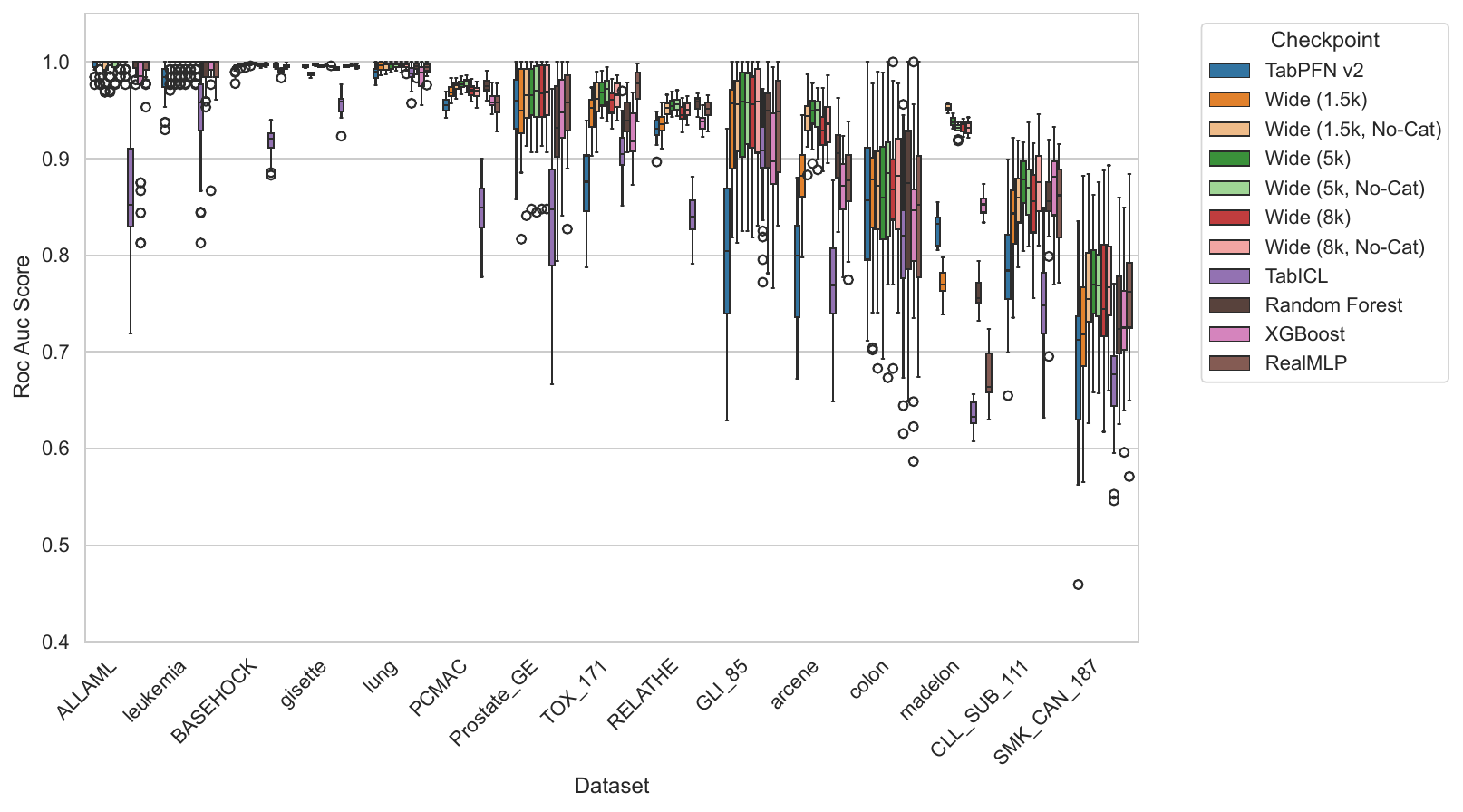}
        \\ \small (b) ROC-AUC per dataset
    \end{minipage}
    \hfill

    \caption{Box plots visualizing the comparison of TabPFNWide in its different variants and other models in terms of Accuracy and ROC-AUC.}
\end{figure}

\subsubsection*{Wilcoxon signed-rank test}
We applied a paired Wilcoxon signed-rank test on the auc-performances of the results shown in~\cref{fig:openml_additional}. The p-values can be inspected in~\cref{tab:wilcoxon_results}.
\begin{table}[htbp]
    \centering
    \caption{Exact p-values from the paired Wilcoxon signed-rank test comparing TabPFN-Wide (5k) against all evaluated baselines across the 15 HDLSS datasets.}
    \label{tab:wilcoxon_results}
    \begin{tabular}{lc}
        \toprule
        \textbf{Model Comparison} & \textbf{p-value} \\
        \midrule
        TabPFN-Wide (5k) vs. Random Forest & $0.00427$ \\
        TabPFN-Wide (5k) vs. RealMLP       & $0.00116$ \\
        TabPFN-Wide (5k) vs. XGBoost       & $0.00012$ \\
        TabPFN-Wide (5k) vs. TabPFN v2     & $0.00018$ \\
        TabPFN-Wide (5k) vs. TabICL        & $6.10 \times 10^{-5}$ \\
        \bottomrule
    \end{tabular}
\end{table}

\section*{Appendix I: Training of TabICL with HDLSS prior}
\label{subsec:tabicl_training}
We tried training TabICL~\citep{qu2025tabicl} with the same training setup as for TabPFN-Wide. However, the model’s training performance did not improve, suggesting that our HDLSS prior may not be effective for TabICL. Whether this arises from TabICL's architectural setup which could make it unsuitable for HDLSS data in general or whether changes to the prior / continued pre-training could mitigate this problem, remains open for future research.  
\begin{figure}[H]
    \centering
    \begin{minipage}[b]{0.45\textwidth}
        \centering
        \includegraphics[width=\linewidth]{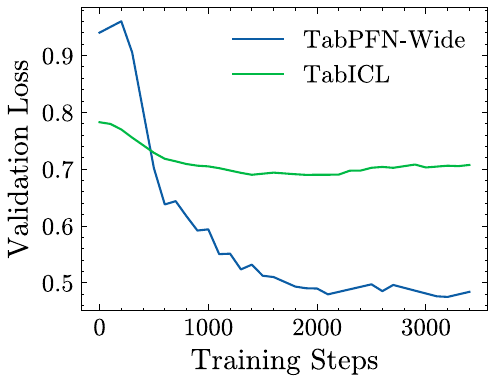}
    \end{minipage}
    \hfill
    \begin{minipage}[b]{0.45\textwidth}
        \centering
        \includegraphics[width=\linewidth]{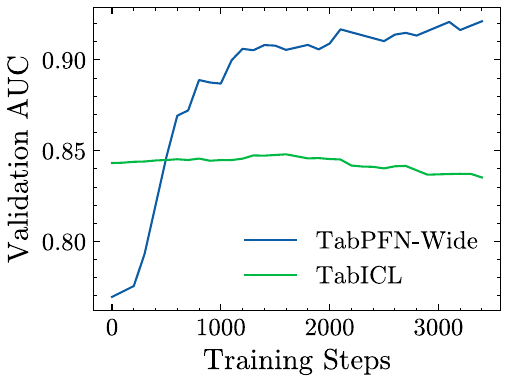}
    \end{minipage}

    \caption{Development of validation loss (left) and validation AUROC (right) for TabICL vs. TabPFN-Wide when training with the same HDLSS prior.}
    \label{fig:train_tabicl}
\end{figure}

\newpage



\section*{Appendix J: Monitoring Model Performance}
\label{subsec:model_selection_monitoring}
To determine an appropriate stopping point for training, we monitored model performance on a set of HDLSS datasets. Specifically, we evaluated performance on two omics datasets and three synthetically generated SNP datasets created with HAPNEST \citep{HAPNEST}. 

\begin{figure}[H]
    \centering
    \begin{minipage}[b]{0.45\textwidth}
        \centering
        \includegraphics[width=\linewidth]{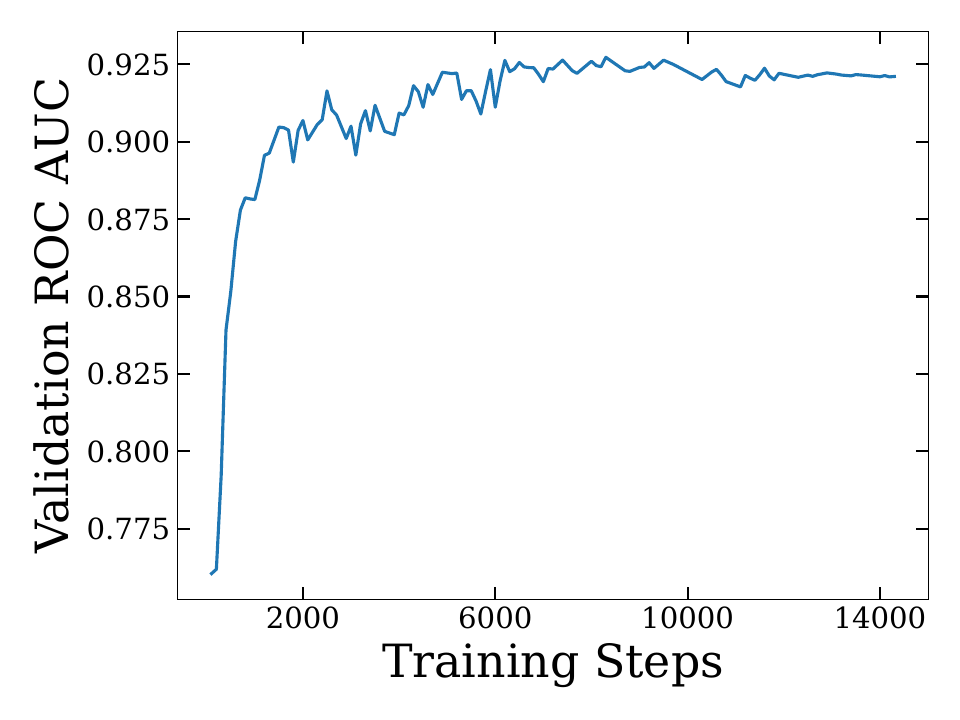}
    \end{minipage}
    \hfill
    \begin{minipage}[b]{0.45\textwidth}
        \centering
        \includegraphics[width=\linewidth]{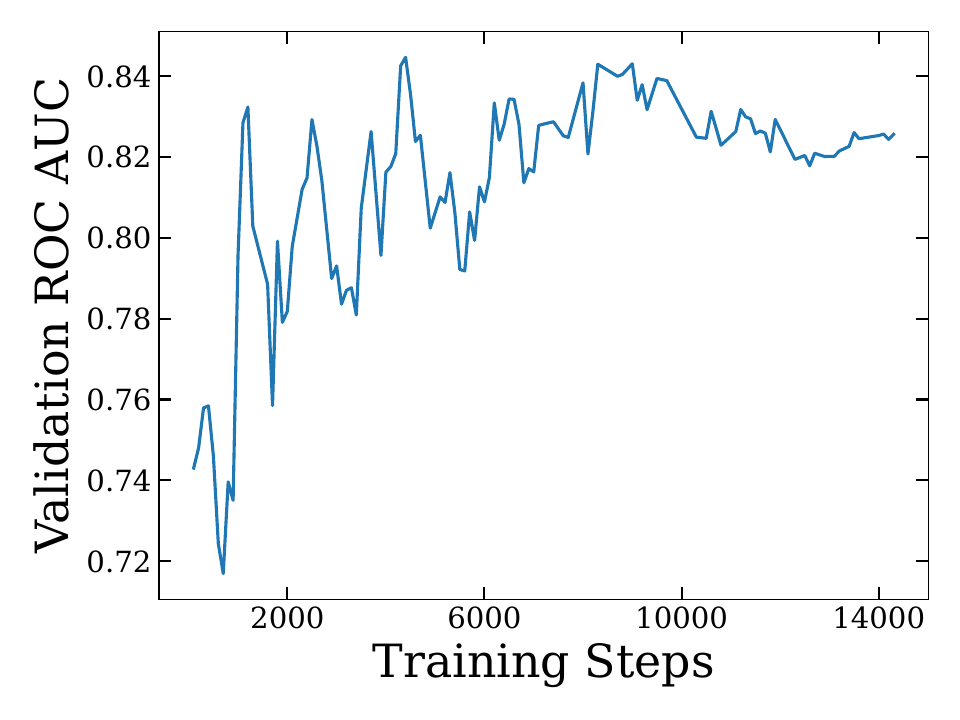}
    \end{minipage}

    \caption{Validation ROC–AUC over training steps for two omics datasets (left) and three HAPNEST-generated SNP datasets (right). Performance plateaus after approximately 10{,}000 steps.}

    \label{fig:train_tabpfn}
\end{figure}

As shown in \cref{fig:train_tabpfn}, validation ROC–AUC improved during the early phase of training but plateaued after approximately 10{,}000 optimization steps. Beyond this point, no consistent performance gains were observed across the monitored datasets. Based on this behavior, we fixed the total training duration to 10{,}000 steps for all models.

Importantly, these datasets were used exclusively for monitoring purposes. They were not involved in gradient updates, hyperparameter tuning, or checkpoint selection.

\newpage
\section*{Appendix K: HAPNEST SNP Simulation Details}
\label{subsec:hapnest_simulation}
\begin{figure}[H]
    \centering
    \begin{minipage}[b]{0.44\textwidth}
        \centering
        \includegraphics[width=\linewidth]{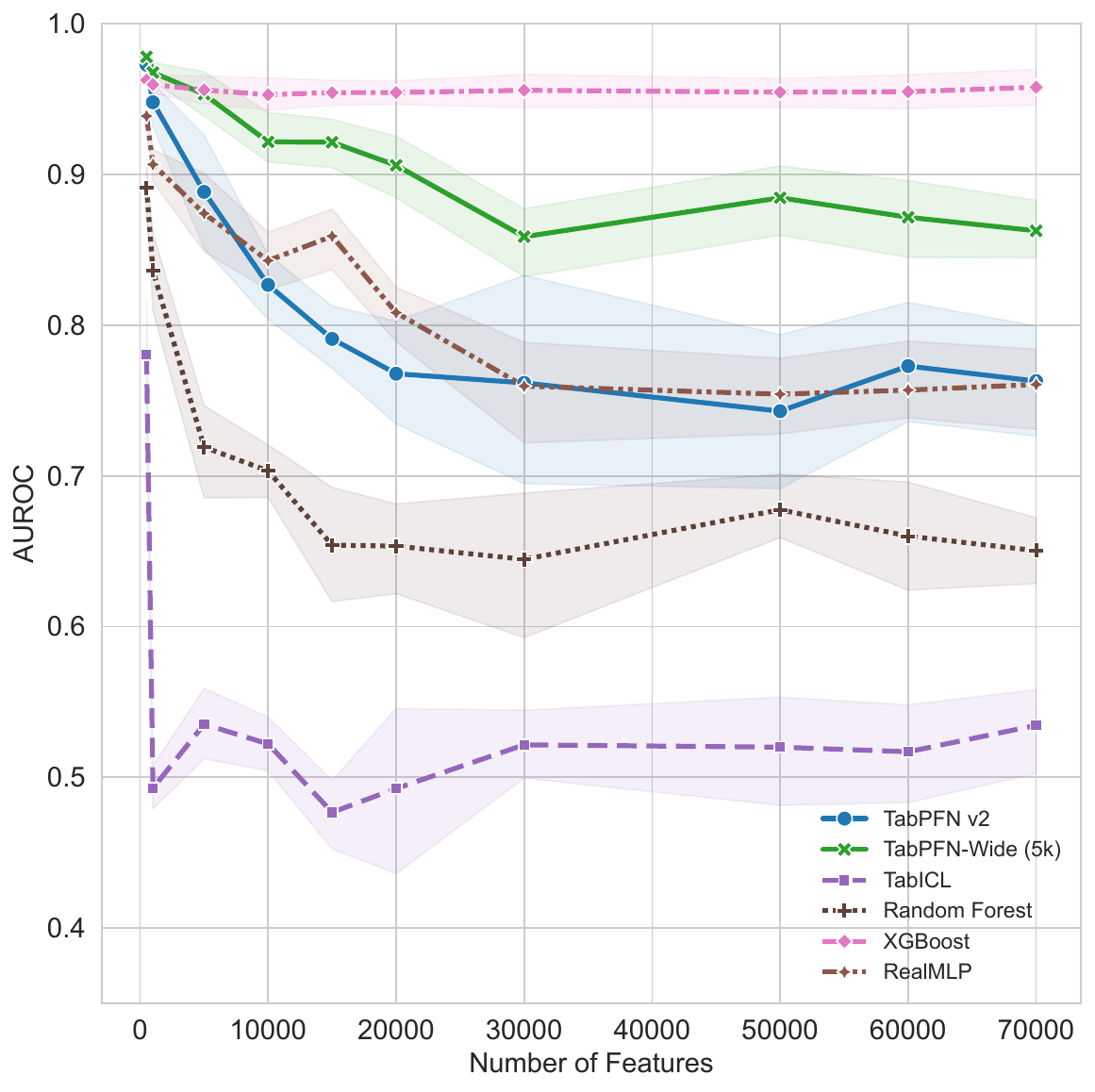}
    \end{minipage}
    \hfill
    \begin{minipage}[b]{0.44\textwidth}
        \centering
        \includegraphics[width=\linewidth]{figures/comparison_plots/reduced_plots/snp_roc_auc_polygenicity_0.01_reduced.pdf}
    \end{minipage}

    \begin{minipage}[b]{0.44\textwidth}
        \centering
        \includegraphics[width=\linewidth]{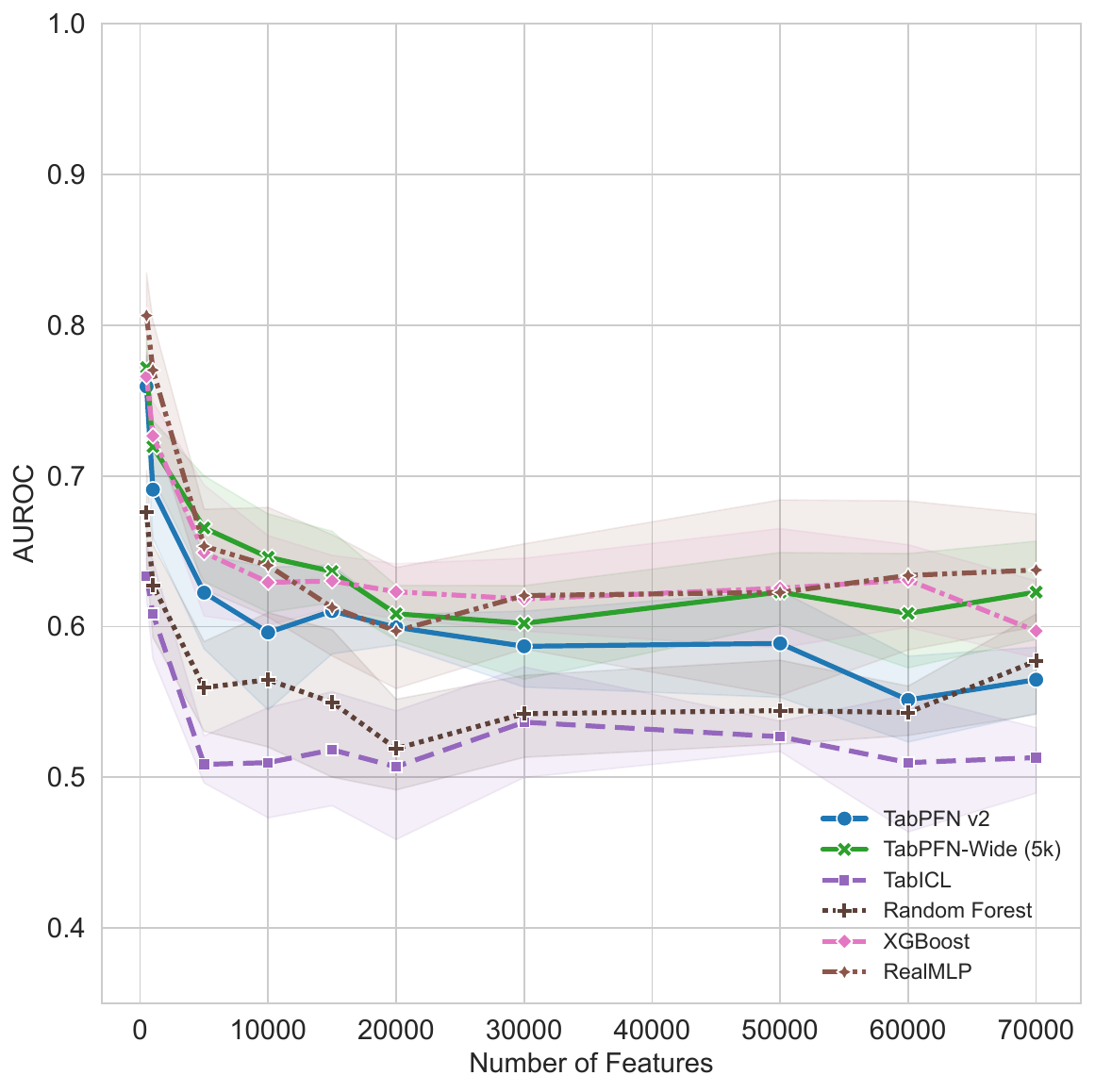}
    \end{minipage}

    \caption{Average AUROC for the SNP datasets with polygenicity of 0.001 (a), 0.01 (b) and 0.05 (c) (higher is better). We compare TabPFN-Wide, using up to 5k features for continued pre-training to TabPFNv2 and other baselines.}

    \label{fig:snp_dataset}
\end{figure}

For the needle-in-a-haystack noise-filtering task presented in the main text, we utilized HAPNEST \citep{HAPNEST} to generate synthetic single nucleotide polymorphism (SNP) datasets. Specifically, we simulated genotypes corresponding to human chromosome~1, which contains on the order of $10^5$ SNPs.\\
Binary phenotypes were generated under a polygenic model where only a small, predefined fraction of the SNPs---referred to as the \textit{polygenicity}---are truly causal for the simulated trait. We evaluated three distinct polygenicity levels: $0.001$, $0.01$, and $0.05$ as shown in~\cref{fig:snp_dataset}. Because chromosome~1 contains roughly $10^5$ SNPs, a polygenicity of $0.001$, for example, results in approximately eight causal SNPs out of the total feature pool.\\
To systematically construct the high-dimensional, low-signal regime, we fixed the set of causal variants for each polygenicity level and progressively introduced non-causal SNPs sampled from the remaining variants on chromosome~1. This controlled approach allowed us to isolate the models' robustness to increasing feature dimensionality and extreme signal sparsity without altering the underlying predictive signal.\\

\newpage

\section*{Appendix L: Feature Correlation Maps}
\label{app:prior_correlation}

As described in the main text, our feature widening procedure induces structured correlation patterns among the generated features because new features only depend on a subset of the original features. The sparsity parameter $p$ controls this structure: small values yield new features influenced by few or no originals, resulting in sparse correlation patterns, whereas large values produce new features that are mixtures of many originals, leading to dense correlation patterns. 

\begin{figure*}[ht]
    \centering
    \begin{tabular}{@{\hskip0cm}c@{\hskip0cm}c@{\hskip0cm}c@{\hskip0cm}c@{\hskip0cm}c@{\hskip0cm}c@{\hskip0cm}c@{\hskip0cm}}
    \centering
     real-world & $p=0$ & $p=0.01$ & $p=0.02$ & $p=0.1$ & $p=1.0$ \\
     \includegraphics[width=0.155\textwidth]{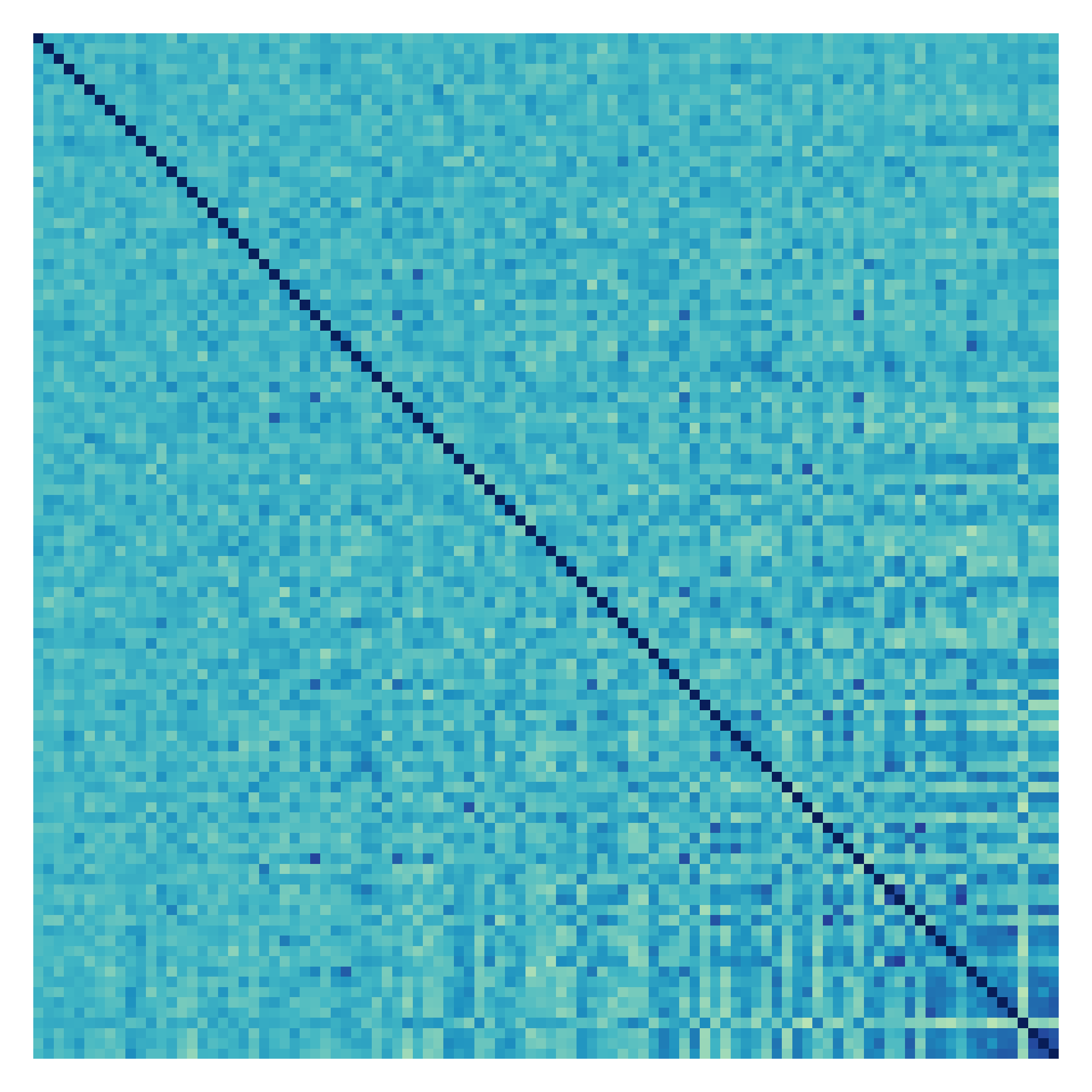} &
     \includegraphics[width=0.155\textwidth]{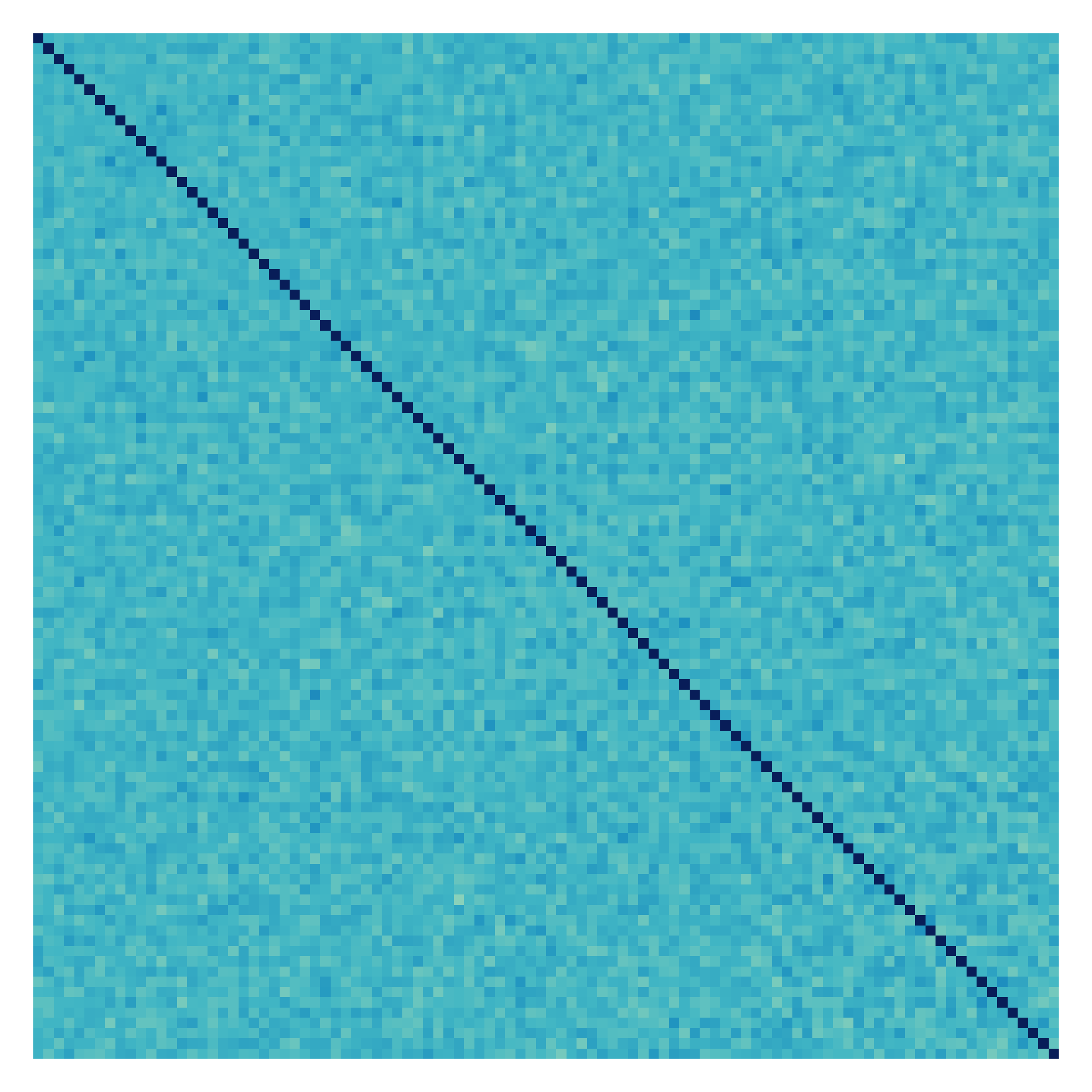} &
     \includegraphics[width=0.155\textwidth]{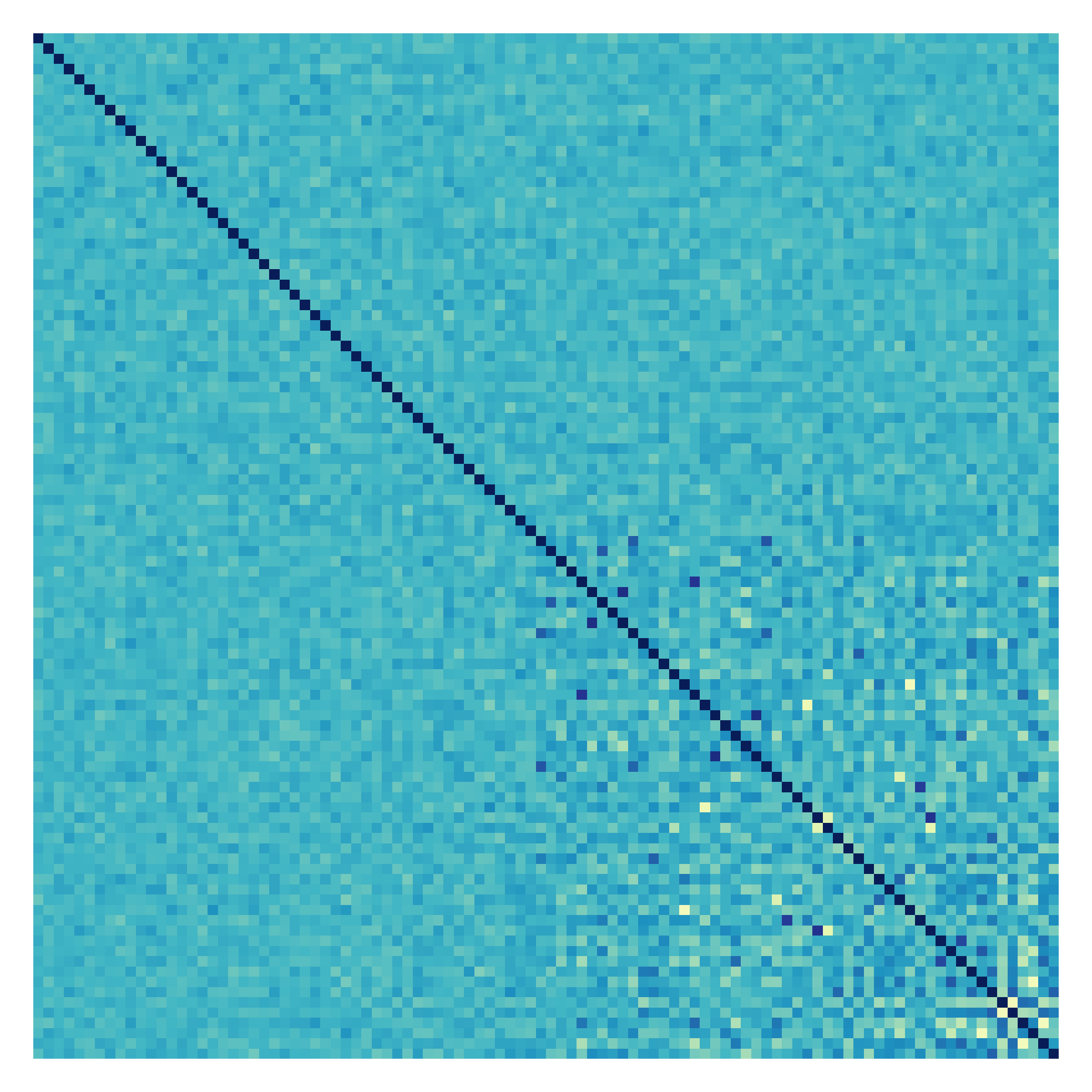} &
     \includegraphics[width=0.155\textwidth]{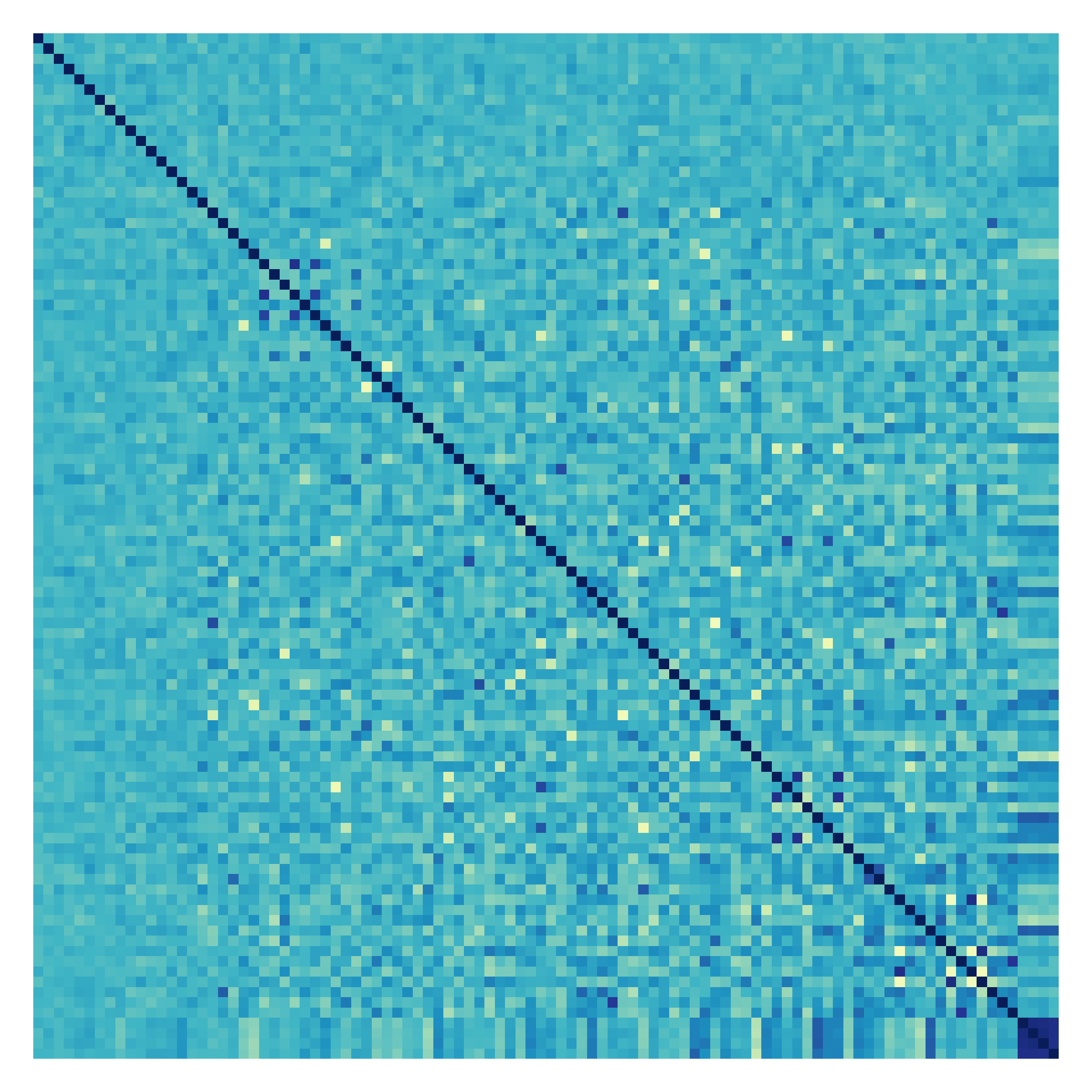} &
     \includegraphics[width=0.155\textwidth]{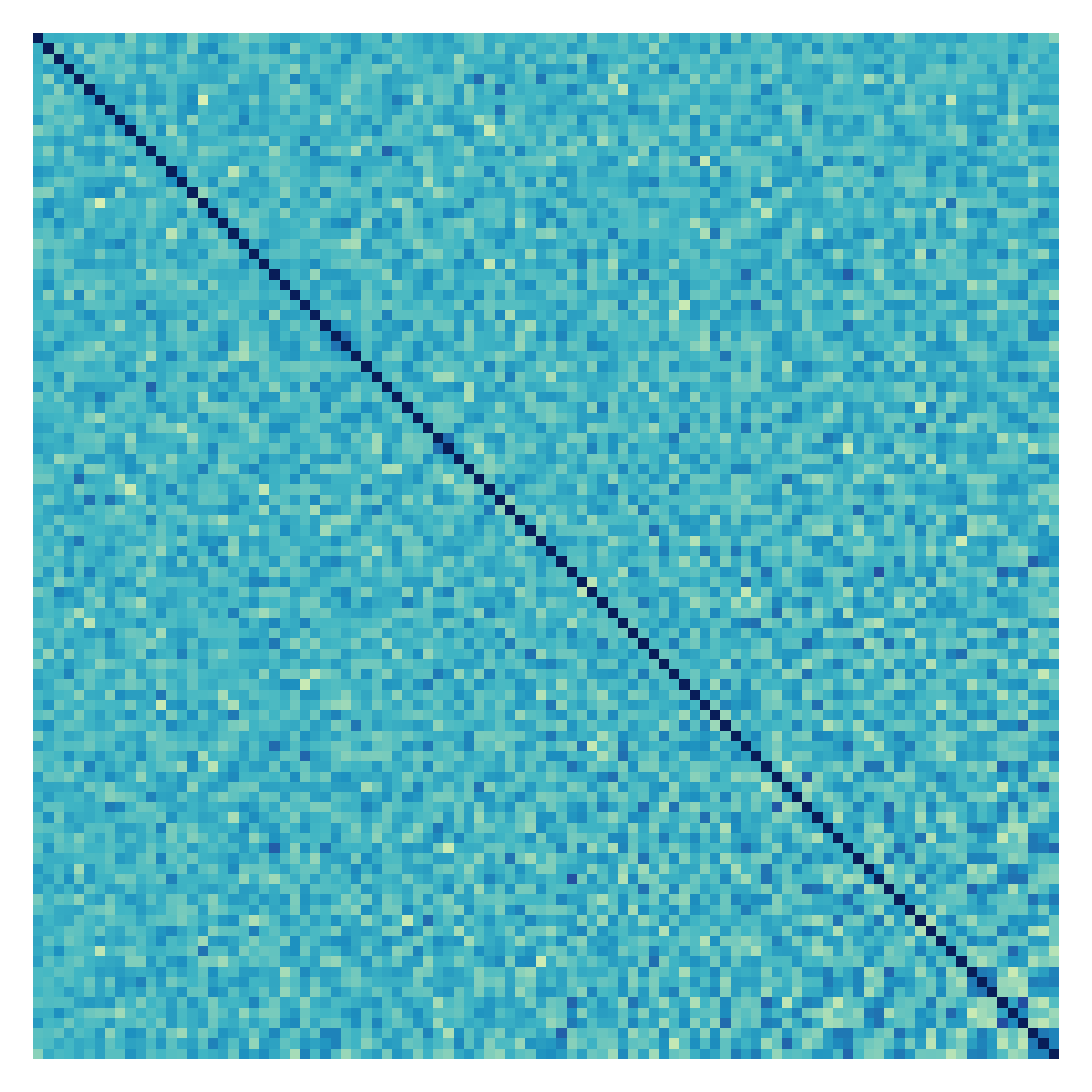} &
     \includegraphics[width=0.155\textwidth]{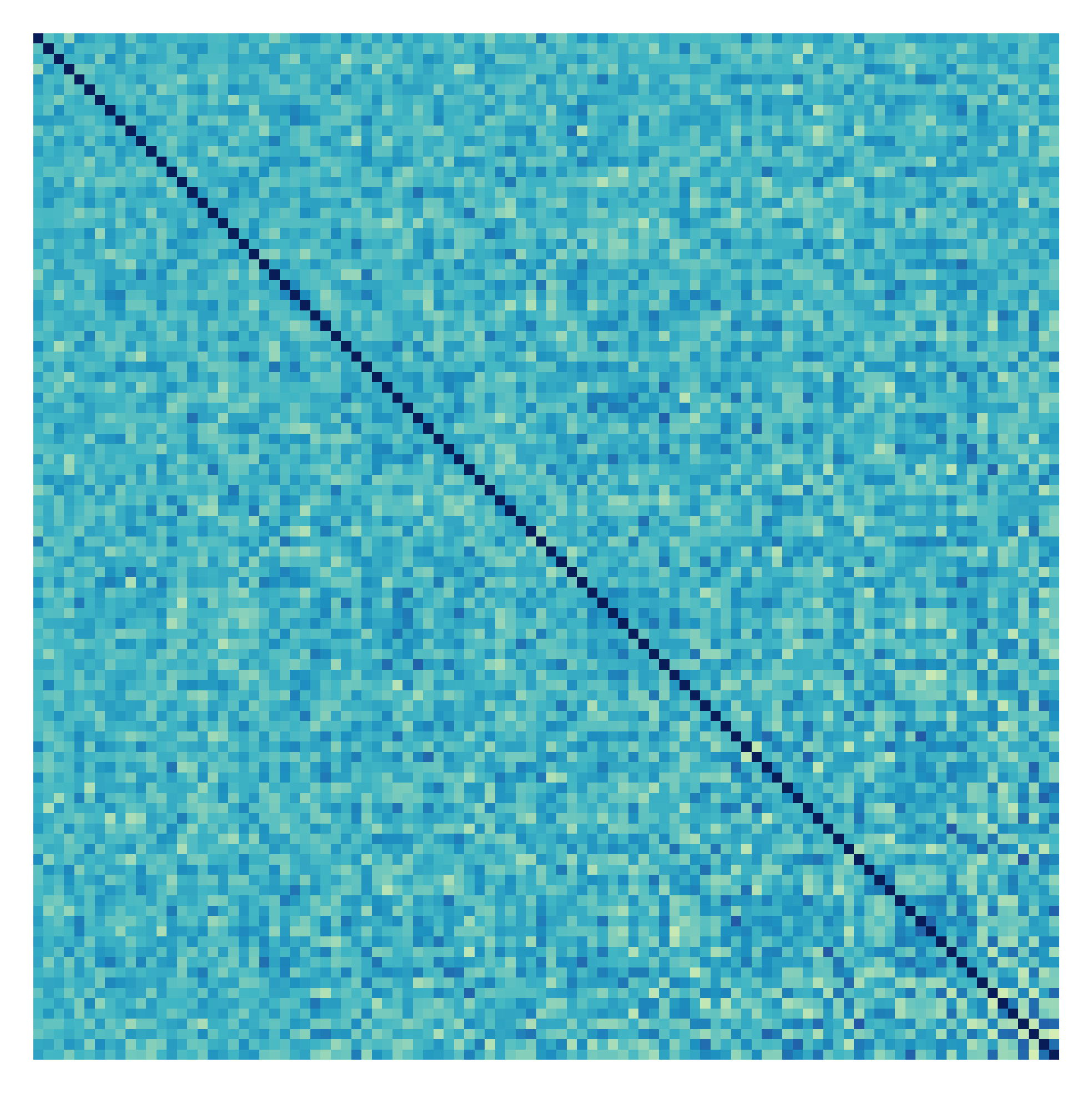} &
     \includegraphics[height=0.155\textwidth]{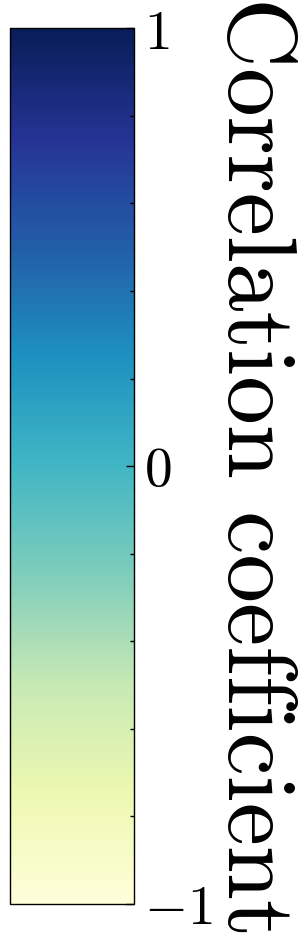} \\
     (a) & (b) & (c) & (d) & (e) & (f) \\
    \end{tabular}
    \caption{Feature correlation maps for (a) mRNA gene expression data and (b-f) synthetically generated datasets with different sparsity values $p$. We compute Pearson correlation for $100$ randomly sampled features and sort them by average absolute correlation.}
    \label{fig:prior_correlation}
\end{figure*}

As an example for continuous features, \cref{fig:prior_correlation} compares real-world HDLSS biomedical data (a) with synthetic datasets (b-f) generated using varying sparsity values. We observe that setting $p=0.02$ shows the closest match to the real correlation structure.
\putbib[reference]
\end{bibunit}

\end{document}